\documentclass[10pt, oneside]{article}

\usepackage{geometry}
\usepackage{amsmath,amsfonts,bm}
\usepackage{wrapfig,lipsum}
\usepackage{natbib}
\usepackage{caption}
\usepackage[utf8]{inputenc}
\usepackage[english]{babel}
\usepackage{amssymb,amsmath,amsthm}

\usepackage{xcolor}
\definecolor{prune}{rgb}{0.44, 0.11, 0.11}
\definecolor{myblue}{rgb}{0, .5, 1}
\definecolor{maroon}{rgb}{0.5450, 0, 0}
\definecolor{darkred}{rgb}{0.5450, 0, 0}
\definecolor{RoyalBlue}{RGB}{0,100,170}
\definecolor{DarkBlue}{RGB}{20,70,200}
\definecolor{peach}{rgb}{1, 0.56, 0.56}
\definecolor{NotionGreen}{RGB}{15,123,108}
\definecolor{NotionOrange}{RGB}{217,115,13}
\definecolor{NotionRed}{RGB}{224,62,62}
\definecolor{red}{RGB}{224,62,62}
\definecolor{midgray}{RGB}{150,150,150}
\definecolor{lavender}{rgb}{0.75, 0.58, 0.89}
\definecolor{Indigo7}{RGB}{66, 99, 235}
\definecolor{Green7}{RGB}{55, 178, 77}
\definecolor{Yellow7}{RGB}{245, 159, 0}
\definecolor{Red7}{RGB}{240, 62, 62}

\def\eqref#1{equation~\ref{#1}}
\def\1{\bm{1}}

\DeclareMathAlphabet{\mathsfit}{\encodingdefault}{\sfdefault}{m}{sl}
\SetMathAlphabet{\mathsfit}{bold}{\encodingdefault}{\sfdefault}{bx}{n}

\theoremstyle{plain}

\theoremstyle{remark}

\usepackage[utf8]{inputenc} % allow utf-8 input
\usepackage[T1]{fontenc}    % use 8-bit T1 fonts

\definecolor{lightblue}{HTML}{2970CC}
\definecolor{lightpurple}{HTML}{673147}
\definecolor{ForestGreen}{HTML}{FF5733}
\definecolor{myred}{HTML}{AA4A44}

\usepackage[colorlinks=true,linkcolor=lightblue,urlcolor=lightblue,citecolor=lightblue,breaklinks]{hyperref}
\usepackage{url}            % simple URL typesetting
\usepackage{booktabs}       % professional-quality tables
\usepackage{amsfonts}       % blackboard math symbols
\usepackage{nicefrac}       % compact symbols for 1/2, etc.
\usepackage{microtype}      % microtypography
\usepackage{xcolor}         % colors
\usepackage{graphicx}
\usepackage{subcaption} 
\usepackage{amsmath,amsfonts,amsthm,amssymb,mathrsfs,dsfont,bbm,bm}
\usepackage{mathtools}
\usepackage{amsmath,amssymb,amsthm,amsfonts,latexsym,bm,bbm,xspace,graphicx,float,mathtools,mathdots,physics,mathrsfs,dsfont}
\usepackage{braket,subcaption,ellipsis,textcomp,hhline,pifont,combelow,booktabs,csquotes}
\usepackage[section]{placeins} % keep floats inside their section
\usepackage{float}             % enables [H] exact placement
\usepackage[font=small,labelfont=bf]{caption}

\usepackage{wrapfig}
\usepackage{enumitem}
\usepackage{authblk}

\usepackage{fancyhdr}
\makeatletter
\newcommand{\samethanks}[1][\value{footnote}]{\footnotemark[#1]}
\makeatother

\usepackage{bm}
\def\*#1{\mathbf{#1}}

\newif\ifshowagcomments
\showagcommentstrue  % Use \showagcommentsfalse to hide comments

\title{Pre-Generating Multi-Difficulty PDE Data\\for Few-Shot Neural PDE Solvers}

\author[1]{Naman Choudhary\thanks{Equal contribution. Author order determined alphabetically.}}
\author[1]{Vedant Singh\samethanks}
\author[1]{Ameet Talwalkar}
\author[1]{Nicholas Matthew Boffi}
\author[2]{\\Mikhail Khodak\thanks{Equal advising.}}
\author[3]{Tanya Marwah\samethanks}

\affil[1]{Machine Learning Department, Carnegie Mellon University}
\affil[2]{Department of Computer Sciences, UW-Madison}
\affil[3]{Simons Foundation}

\date{}

\begin{document}

\maketitle

% !TEX root = main.tex

\begin{abstract}
A key aspect of learned partial differential equation (PDE) solvers is that the main cost often comes from \emph{generating} training data with classical solvers rather than learning the model itself. 
Another is that there are clear \emph{axes of difficulty}---e.g., more complex geometries and higher Reynolds numbers---along which problems become (1)~harder for classical solvers and thus (2)~more likely to benefit from neural speedups. 
Towards addressing this chicken-and-egg challenge, we study \emph{difficulty transfer} on 2D incompressible Navier-Stokes, systematically varying task complexity along geometry~(number and placement of obstacles), physics~(Reynolds number), and their combination. 
Similar to how it is possible to spend compute to {\em pre-train} foundation models and improve their performance on downstream tasks, we find that by classically solving~(analogously {\em pre-generating}) many low and medium difficulty examples and including them in the training set, it is possible to learn high-difficulty physics from far fewer samples.
% Furthermore, we show that, for a fixed compute budget, including \emph{medium} difficulty data outperforms using easy data, requires fewer overall samples, and achieves the same error with up to $5\times$ fewer hard examples.
% Furthermore, we show that using compute on low and medium difficulty
% examples 
% training data 
Furthermore, we show that by combining low and high difficulty data, we can spend 8.9$\times$ less compute on pre-generating a dataset to achieve the same error as using only high difficulty examples.
Our results highlight that \emph{how} we allocate classical-solver compute across difficulty levels is as important as \emph{how much} we allocate overall, and suggest substantial gains from principled curation of pre-generated PDE data for neural solvers. Our code is available \href{https://github.com/Naman-Choudhary-AI-ML/pregenerating-pde}{at this url}.\footnotemark

\footnotetext{Code: \url{https://github.com/Naman-Choudhary-AI-ML/pregenerating-pde}}
\end{abstract}

% !TEX root = main.tex

%\vspace{-3mm}
\section{Introduction}
%\vspace{-2mm}

Deep learning has emerged as a powerful paradigm for solving PDEs, enabling data-driven surrogate models that can accelerate simulation, inference, and design across diverse scientific domains~\citep{li2021fourierneuraloperatorparametric, Lu_2021, pathak2022fourcastnetglobaldatadrivenhighresolution}.
This has been driven by the development of specialized models such neural operators~\citep{li2021fourierneuraloperatorparametric,Lu_2021} and recent transformer-based extensions~\citep{guibas2022adaptivefourierneuraloperators, brandstetter2023messagepassingneuralpde}, which have demonstrated strong performance on benchmark datasets and gained traction in the machine learning for science community.
More recently, significant effort has been devoted towards pre-generation of large datasets such as the Well~\citep{ohana2024well}
and pre-training specialized 
foundation models~(FMs)~\citep{herde2024poseidon, hao2024dpot, shen2024ups, mccabe2023multiple}.
The goal of these FMs is to serve as general-purpose foundations for PDE surrogates:
delivering fast inference while minimizing---or even eliminating---the need to retrain on new, potentially harder-to-solve PDEs.
% The latter is critical for making neural solvers useful for applications such as inverse design and optimization~\citep{li2022mlaso,shukla2023deepoperator,glaws2022inn,ma2020dlphotonics,tahersima2019nanophotonic}.\tm{TODO: naman and vedant, add citations here. (Misha edit: specifically inverse problems, inverse design, shape optimization, computer-assisted engineering, etc.)}

% {\color{red}
% Many inverse design and PDE-constrained optimization tasks require repeated solves across configurations, making classical loops costly (e.g., aerodynamic shape design, nanophotonics)~\citep{li2022mlaso}. Neural surrogates alleviate this by delivering near-instant PDE evaluations—DeepONet surrogates cut online airfoil-optimization cost by orders of magnitude~\citep{shukla2023deepoperator} (see also \citep{glaws2022inn}), and end-to-end nanophotonic inverse design achieves order-of-magnitude speedups~\citep{ma2020dlphotonics,tahersima2019nanophotonic}. For inverse problems, physics-informed neural networks offer efficient alternatives to classical iterative methods~\citep{raissi2019pinn,karniadakis2021piml}. Collectively, these advances accelerate CAE workflows—shortening design iterations, optimization loops, and enabling near-real-time decision support.
% }

An underlying aspect of this line of work has long been the issue that it seeks to solve PDEs faster than classical numerical solvers but requires examples generated by the latter to do so.
% %
% \nb{Maybe something like -- many of the engineering tasks we may hope AI can accelerate? Also, can we give some specific examples of this?}
% %
While such a circuitous setup is justifiable in many of the inverse problem applications that motivate learned solvers, it is still the case that tasks we eventually want to accelerate---practical engineering tasks in difficult-to-simulate regimes---will be exactly those for which it is hard to generate a significant amount.
% %
% \nb{Why is it that surrogates would be useful for these hard regimes? Is it because we expect them to be more accurate? Is it because we want to use them in some kind of outer-loop application? If I \textit{can} solve the equation with a classical technique (which our assumption is that you can -- because we need the data), then why do I even want a surrogate in the first place? I don't think the role of a surrogate or why they're useful has really been explained in the intro here. We should connect to inverse problems, inverse design, shape optimization, etc.}
% %
% Precisely for the hardest regimes where surrogates would be most useful, classical simulation is slow and expensive, making training data scarce.
This need to decrease the sample complexity of neural PDE solvers has spurred significant research drawing up transfer learning~\citep{herde2024poseidon}, active learning~\citep{bruna2024neural,musekamp2024active}, and other {\em method-centric} approaches~\citep{rotman2023semi}.\looseness-1

In this paper we take a {\em data-centric} view, studying how the training data composition of neural PDE solvers affects their performance.
We identify that a key feature of PDE data is that most problem settings have multiple axes of difficulty along which classical solving becomes harder, thus making neural PDE solvers both (potentially)~more useful but also more difficult to train due to low data availability.
Examples of such difficulty axes include domain geometry features, physics parameters such as the Reynolds number~(Re) or Debye length, additional terms due to forcing or compressibility, and so on.
To understand how easier-to-generate data affects the performance on harder-to-generate target distributions, we consider incompressible Navier-Stokes simulations with difficulty varying along either or both of (1)~{\em geometry}---as defined by the number and complexity of obstacles in the flow---or (2)~{\em physics} in terms of the flow's Re.
For simplicity, we use classical simulation costs to divide these two axes into three difficulty categories---easy, medium, and hard---and investigate how mixing in easy and medium data affects performance on the hard distribution.

Our first key result is that {\bf adding easy-to-medium difficulty data substantially improves performance} on the hard distribution.
Naturally, one might expect that medium difficulty data might be more useful, and our second main result is that there is often {\bf a favorable tradeoff justifying pre-generating medium-difficulty datasets} instead of easy ones when solving cross in two classes of simulation, flow past an object~(FPO) and lid-driven cavity~(LDC), and using both supervised-only neural PDE solvers---specifically the Factorized Fourier Neural Operator~(FFNO)~\citep{tran2021factorized} 
and the Convolutional Neural Operator~(CNO)~\citep{raonic2023convolutional}---and the current state-of-the-art multi-physics pretrained Poseidon FM~\citep{herde2024poseidon}.
These complementary settings allow us to assess both specialized neural operators and large pretrained models under controlled difficulty-mixing regimes.
In more detail, our contributions are the following:
% Naturally, one might expect that simulation data in the medium difficulty range may be more useful for learning hard distributions than data from low difficulty ranges.
% Our key finding is that---at least for 2D fluids---there is a favorable tradeoff justifying pre-generating datasets of medium difficulty data in order to solve higher difficulty tasks.
\begin{enumerate}[leftmargin=*,topsep=0mm,itemsep=1pt,parsep=0pt]
    \item 
    Augmenting hard (e.g., multi‑obstacle) training with lower‑difficulty data (e.g., zero or one obstacle) substantially improves accuracy on the hard test set.
    For example, most of the performance of Poseidon-B fine-tuned solely on hard FPO data can be recovered when 90\% of it is replaced with easy or medium data , which reduces data-generation time 8.9$\times$.\looseness-1
    \item Despite the higher generation cost of medium difficulty (e.g., single-obstacle) examples relative to easy (e.g., no obstacle) ones, for most pre-generation budgets one can obtain a better error by training on fewer examples of the former rather than more of the latter.
    This demonstrates the importance of optimally selecting the pre-generation simulations.
    \looseness-1
    \item Beyond square obstacles, we show that single simple‑obstacle data can improve the few-shot performance of models on flows around more complex shapes from FlowBench~\citep{tali2024flowbenchlargescalebenchmark}, even when given only a few examples from it.
    This demonstrates the potential of a single dataset serving as a foundation for few-shot training of learned solvers on {\em multiple} harder datasets.\looseness-1
\end{enumerate}

We have released all pre-generated datasets and code to reproduce our results \href{https://github.com/Naman-Choudhary-AI-ML/pregenerating-pde}{at this url}.
For related work, please see Appendix~\ref{sec:related}.

\section{Pre-generating datasets for studying difficulty transfer}\label{sec:setup}
%\vspace{-2mm}

% --- Figure 1: Snapshots + NURBS (CG & CP + FlowBench) ---
As discussed in the introduction, PDE tasks often feature gradations of difficulty that
significantly increase the cost of simulation, making neural PDE solvers both more expensive to train (because of the complexity of generating the associated data) and potentially more useful (because of their ability to replace said expensive solves).
This increased numerical difficulty can stem from shorter timesteps, higher per‑timestep cost
(e.g., worse conditioning of linear solves), and meshing challenges.
To study how low-to-medium difficulty data can improve few-shot performance on high difficulty data, we consider the 2D incompressible Navier-Stokes~(INS) equations of fluid flow.
Given a domain $\Omega\subset[0,1]^2$, they govern the velocity $\*u(\*x,t)$ and pressure $p(\*x,t)$ of a fluid at point $\*x\in\Omega^\circ$ and time $t\ge0$ as follows:
\begin{equation}
    \partial_t\*u+(\*u\cdot\nabla)\*u+\nabla p=\nu\Delta\*u\qquad\textrm{and}\qquad\nabla\cdot\*u=0
\end{equation}
Different simulation settings can be defined using different domains $\Omega$, different boundary conditions $\*u(\*x,t)$ and $\*p(\*x,t)$ for $\*x\in\partial\Omega$, different initial conditions $\*u(\*x,0)$ and $\*p(\*x,0)$ for $\*x\in\Omega^\circ$,  and different kinematic viscosities $\nu\ge0$.
We focus on two canonical settings:
(1)~flow past an object~(FPO), in which the boundary conditions impose two no slip walls (Dirichlet $u = 0$) around an inlet and an outlet, and (2)~lid-driven cavity flow~(LDC), which has three no slip walls and a horizontal velocity at the top;
in both cases the interior of the domain is at rest to start.\looseness-1

\vspace{-2mm}
\subsection{Difficulty axes}\label{sec:axes}
\vspace{-1mm}

\begin{figure}[!t]
    \begin{minipage}{0.72\linewidth}
        \includegraphics[width=\linewidth]{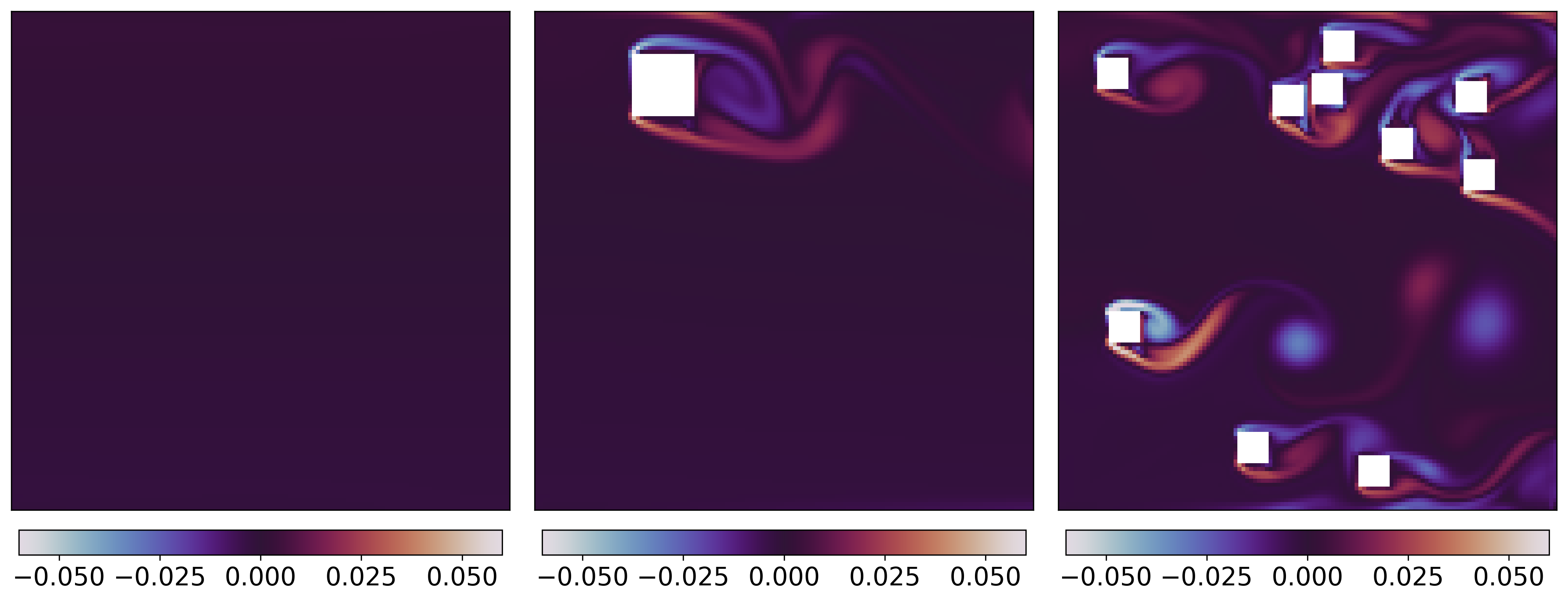}
        \includegraphics[width=\linewidth]{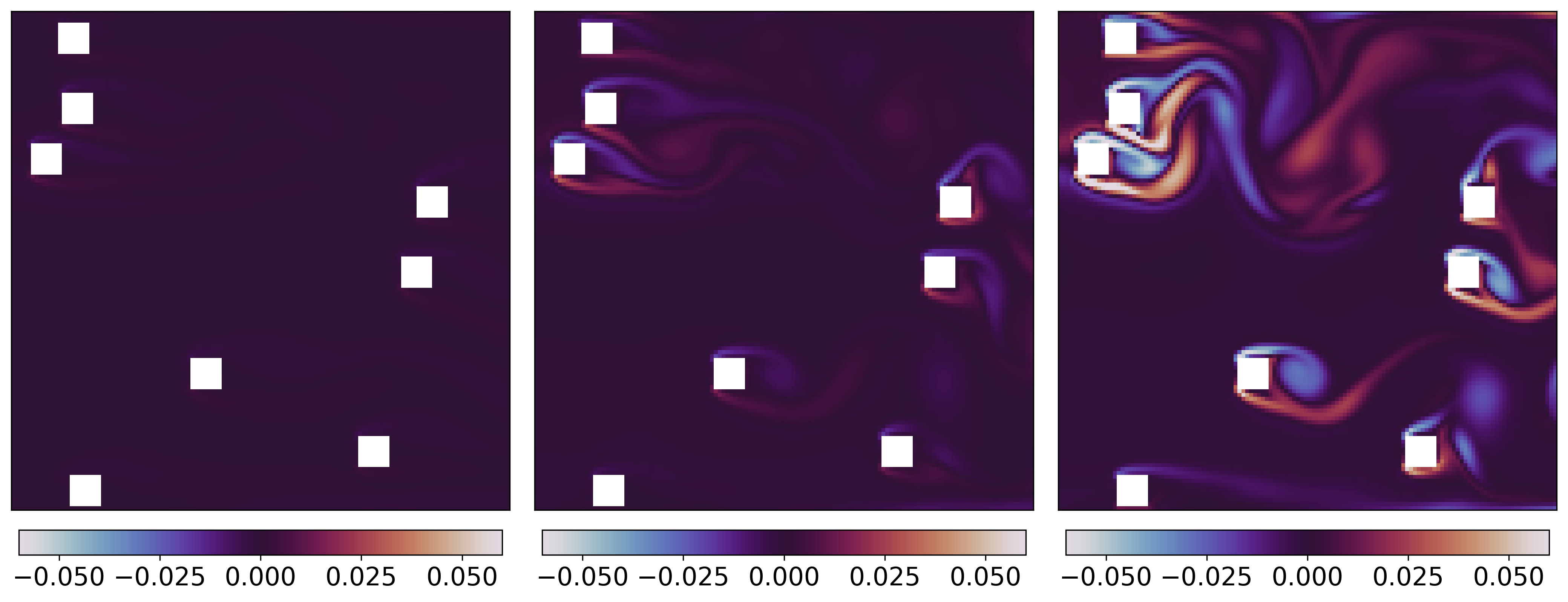}
        \caption{\label{fig:axes-snapshots}
            {\bf Top:} vorticity snapshots across increasing geometry difficulty, with flows past zero, one, and multiple (2–10) square obstacles.
            {\bf Bottom:} snapshots across physics difficulties in the form of low ($[100,1000]$), medium ($[2000,4000]$), and high ($[8000,10000]$) Re bands.\looseness-1
        }
    \end{minipage}
    \hfill
    \begin{minipage}{0.24\linewidth}
        \includegraphics[width=\linewidth]{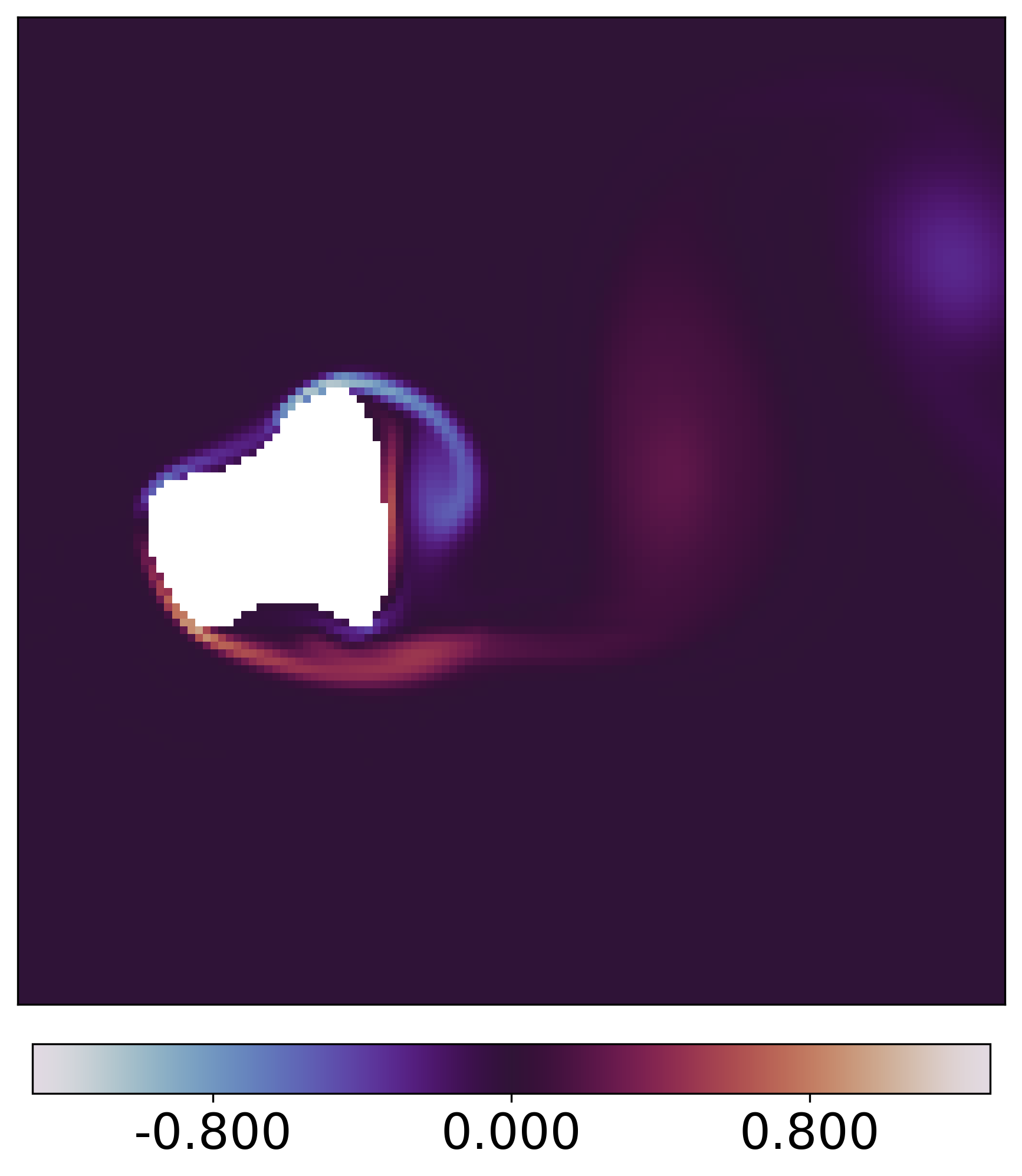}
        \includegraphics[width=\linewidth]{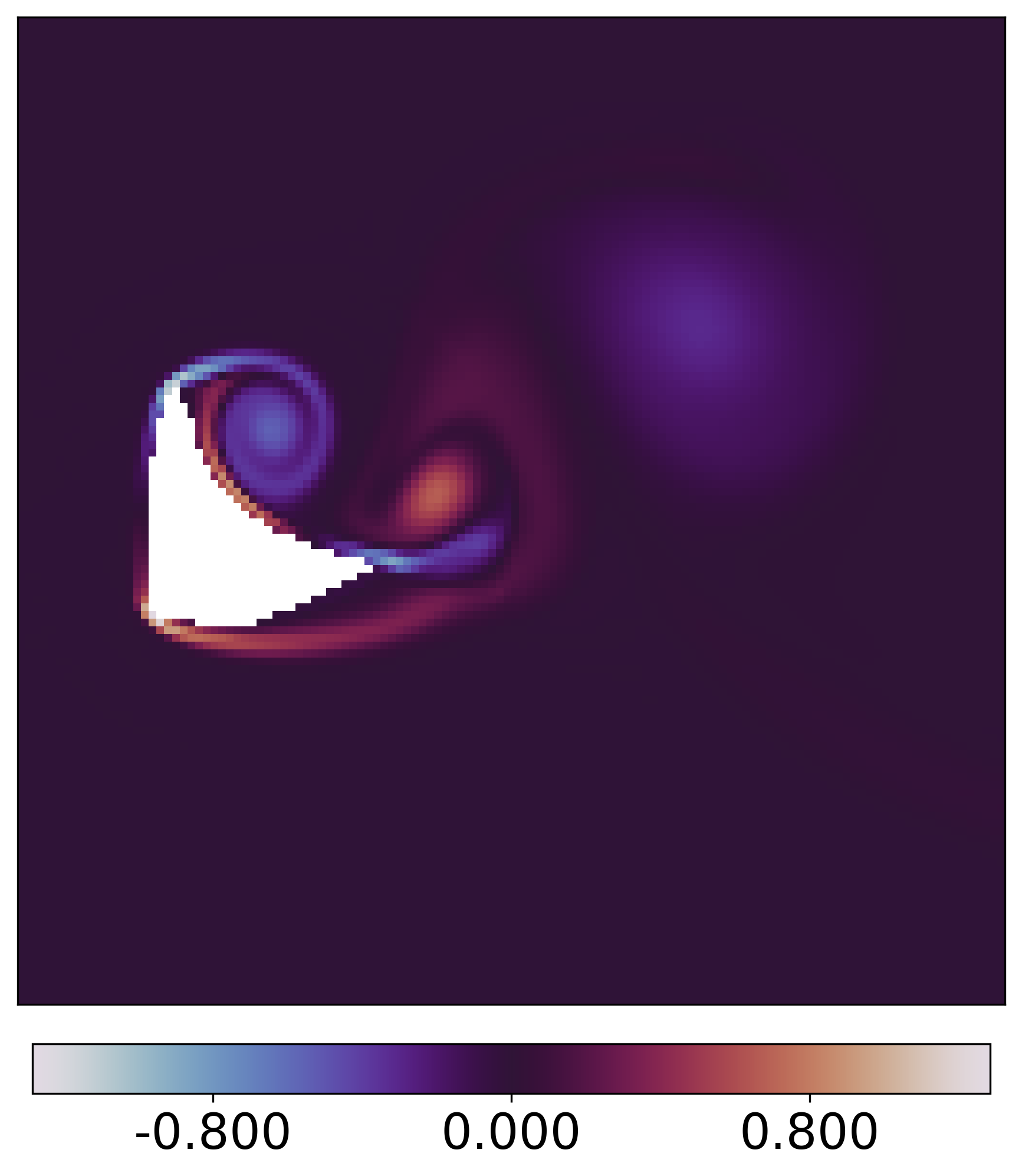}
        \caption{\label{fig:axes-nurbs}
            FPO with objects from the FlowBench G1 NURBS data~\citep{tali2024flowbenchlargescalebenchmark}.
            }
    \end{minipage}
  % \centering
  % % Tweak locally without preamble helpers
  % \def\Gap{0.02\linewidth}
  % \def\LeftW{0.72\linewidth}
  % \def\RightW{0.24\linewidth}
  % \def\RowH{0.28\textheight}
  % % Row 1: Changing Geometry (left) + FlowBench NURBS (right)
  % \begin{minipage}[t]{\LeftW}
  % \end{minipage}\hspace{\Gap}%
  % \begin{minipage}[t]{\RightW}
  % \end{minipage}
  % \vspace{0.2em}
  % % Row 2: Changing Physics (left) + FlowBench NURBS (right)
  % \begin{minipage}[t]{\LeftW}
  %   \par\smallskip\centering\footnotesize\textbf{(a)}
  % \end{minipage}\hspace{\Gap}%
  % \begin{minipage}[t]{\RightW}
  %   \par\smallskip\centering\footnotesize\textbf{(b)}
  % \end{minipage}
  % \caption{\textbf{Difficulty axes for 2D Navier–Stokes.} 
  % \label{fig:axes-snapshots}
  % \vspace{-0.9em} % nudge up next paragraph
\end{figure}

\begin{figure}[!t]
  \centering
  \includegraphics[width=1.0\linewidth]{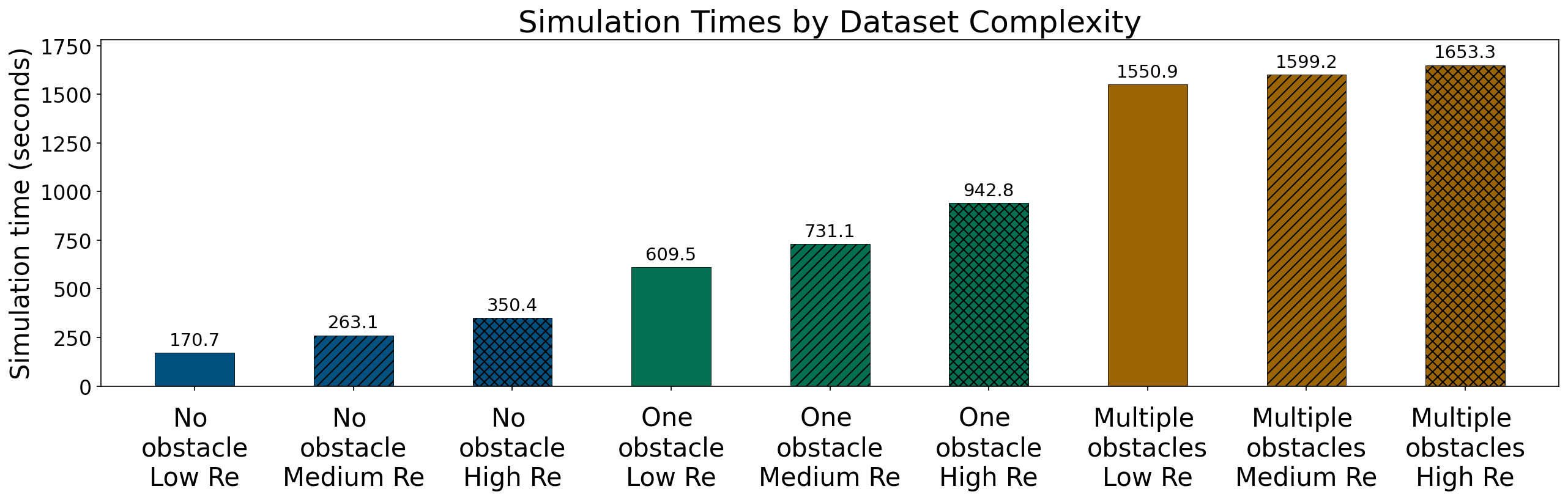}
  \caption{
    {\bf Computational cost} of simulating flow past an object~(FPO) at different difficulty settings, demonstrating increasing runtime along both the domain geometry axis~(increasing number of obstacles) and the physics axis~(increasing Reynolds number).
    The costs reported are averages across thirty simulations.
    }
  \label{fig:simtime-fpo}
  % nudge up next paragraph
\end{figure}

Starting from these basic setups, we vary simulations along three data-difficulty axes: geometry, physics, and their combination.
As detailed below, changing the geometry involves modifying the domain $\Omega$ and its boundary conditions to add, remove, or change the shape of obstacles, with a greater number of objects or more complex shapes corresponding to greater difficulty.
On the other hand, changing the physics involves varying the initial velocity $\*u(\*x,0)$ to change the Reynolds number, a dimensionless quantity that when increased typically makes the flow more complex and hard-to-simulate.
Figures~\ref{fig:axes-snapshots} and~\ref{fig:axes-nurbs} illustrate how the vorticity fields of the simulations change with increasing difficulty, while Figure~\ref{fig:simtime-fpo} shows the corresponding increase in simulation cost.

We next describe at a high level the settings used to generate the axes' data; 
further details, including about our OpenFOAM~\citep{jasak2007openfoam} setup, can be found in Appendix~\ref{app:dataset}.

% \begin{itemize}[leftmargin=*]
%     \item geometry (number/complexity of obstacles),
%     \item physics (Reynolds‑number band),
%     \item both simultaneously.
% \end{itemize}
%
% \nb{What is NURBS?}
%

% Illustrative vorticity fields spanning these axes appear in Fig.~\ref{fig:axes-snapshots}; Non-Uniform Rational B-Splines (NURBS) examples of FlowBench shapes are shown in the right column of the same figure.
% Simulation details are provided in the supplement; we use the OpenFOAM package~\citep{jasak2007openfoam} to 
% generate our datasets.\looseness-1
% \vspace{-0.8em}

% consists of parametric shapes generated using Non-Uniform
% Rational B-Splines (NURBS) curves
% \nb{I wonder if each of these subsections should just be a paragraph environment.}
% First, we describe the settings of the two main difficulty axes, physics and geometry, that we combine to form the third.
% For both the FPO and LDC settings we simulate 6,400 trajectories and save twenty timesteps of velocity and pressure, which the model must predict from the given initial conditions.
\begin{enumerate}[leftmargin=*,topsep=0mm,itemsep=1pt,parsep=0pt]
    \item {\bf Geometry:} 
    A straightforward way to change the problem geometry to increase problem difficulty is by adding or removing obstacles to the flow.
    In this difficulty axis, we add between zero and ten square obstacles at random, non-overlapping positions.
    The resulting simulations are categorized as {\em easy} if they have no obstacles, {\em medium} if they have one obstacle, and {\em hard} if they have two or more.
    The way these changes affect the simulation is illustrated in the top row of Figure~\ref{fig:axes-snapshots}, and their effect on the FPO generation cost is plotted in Figure~\ref{fig:simtime-fpo};
    in short, more obstacles yield more complex, harder-to-simulate flows.\looseness-1
    \item {\bf Physics:}
    Another way of increasing problem difficulty is to increase the Reynolds number, which is well-known to describe the complexity of a flow.
    It is defined using a characteristic velocity $U$ and length-scale $L$ to be Re $=UL/\nu$, so we increase the initial velocity $\*u(\*x,0)$ at the inlet~(FPO) or the lid~(LDC) to make it larger.
    In particular, we categorize simulations into {\em easy}, {\em medium}, and {\em hard} if the corresponding Re is between $[100,1000]$, $[2000,4000]$, and $[8000,10000]$, respectively;
    within each band, the Re is sampled from a truncated Gaussian distribution.
    Figure~\ref{fig:axes-snapshots}~(bottom) demonstrates how a higher Re induces richer fluid structure, yielding the higher simulation cost~(cf. Figure~\ref{fig:simtime-fpo}).
\end{enumerate}

As mentioned before, we combine the geometry and physics axes to obtain our third difficulty axis;
in the latter case we use low Re flows with no objects as the easy examples and medium Re flows with one object as the medium examples. In all cases, we treat “easy/medium/hard” as a relative, cost-based notion of difficulty: configurations that are cheaper to solve (e.g., low-Re, simple geometries) form the easy tier, while those that require substantially more wall-clock time (e.g., high-Re, multi-obstacle flows) form the hard tier (see Figure ~\ref{fig:simtime-fpo}).

% --- Figure 2: Cost (placed right after its first reference) ---

Lastly, we also enhance our investigation of geometry difficulty by changing the object complexity, specifically by incorporating simulations from FlowBench~\citep{tali2024flowbenchlargescalebenchmark}.
Specifically, we use their G1 dataset of FPO simulations of flows around Non-Uniform Rational B-Splines~(NURBS), two examples of which are depicted in Figure~\ref{fig:axes-nurbs}.
Because FlowBench is external, we do not measure the classical solve difficulty and treat no-obstacle and single (square) obstacle flows from the geometry axis as the easy and medium difficulty settings, respectively.\looseness-1

\vspace{-2mm}
\subsection{Pre-generated datasets}
\vspace{-1mm}

For each of the above settings and difficulty axes~(excluding FlowBench) we generate $n=6,400$ simulations with different randomly sampled initial conditions.
Specifically, following \citet{herde2024poseidon} we sample Re between $[100,1000]$ (or higher if we are varying along the physics axis, as described above) and use that to set the inlet~(FPO) or lid~(LDC) velocity.
We store the solution $\bm{y}_t^i = (\*u_i(\*x,t),p_i(\*x,t))$ of each simulation $i=1,\dots,n$ at $T=20$ timesteps $t=1,\dots,T$ on a regular grid of points $\*x\in\Omega$.
Starting with this data, we hold out a subset of $N=100$ trajectories and set the goal of a learned PDE solver as using the remaining data to learn a model that, given the initial conditions 
$\bm{y}_0^i = (\*u_i(\*x,0),p_i(\*x,0))$ 
of a held-out trajectory $i$,
predicts a trajectory $\hat{\bm{y}}_t^i$ where $t \in [1, T]$
and $\hat{\bm{y}}_t^i = (\*{\hat u}_i(\*x,t),\hat p_i(\*x,t))$.
% $\{(\*{\hat u}_i(\*x,t),\hat p_i(\*x,t))\}_{t=1}^T$ such that $(\*{\hat u}_i,\hat p_i)\approx(\*u_i,p_i)$.
Following \citet{raonic2023convolutional,herde2024poseidon}, we measure its success at doing so using the mean relative L1 error~(nMAE):
\begin{equation}
% \color{red}
\mathrm{nMAE}
=\displaystyle \sum_{i=1}^{N} \sum_{t=1}^{T} \left\| \bm{y}_t^{\,i} - \hat{\bm{y}}_t^{\,i} \right\|_{1}\bigg/
\displaystyle \sum_{i=1}^{N} \sum_{t=1}^{T} \left\| \bm{y}_t^{\,i} \right\|_{1}
\label{eq:nmae}
\end{equation}
% Unless otherwise stated, we use nMAE in our main evaluation. %We also report the L2 error in the appendix.

% !TEX root = main.tex

% \section{Empirical results}

%\vspace{-3mm}
\section{Empirical results}\label{sec:experiments}
%\vspace{-2mm}

\begin{figure}[!t]
  \centering
\includegraphics[width=0.495\linewidth]{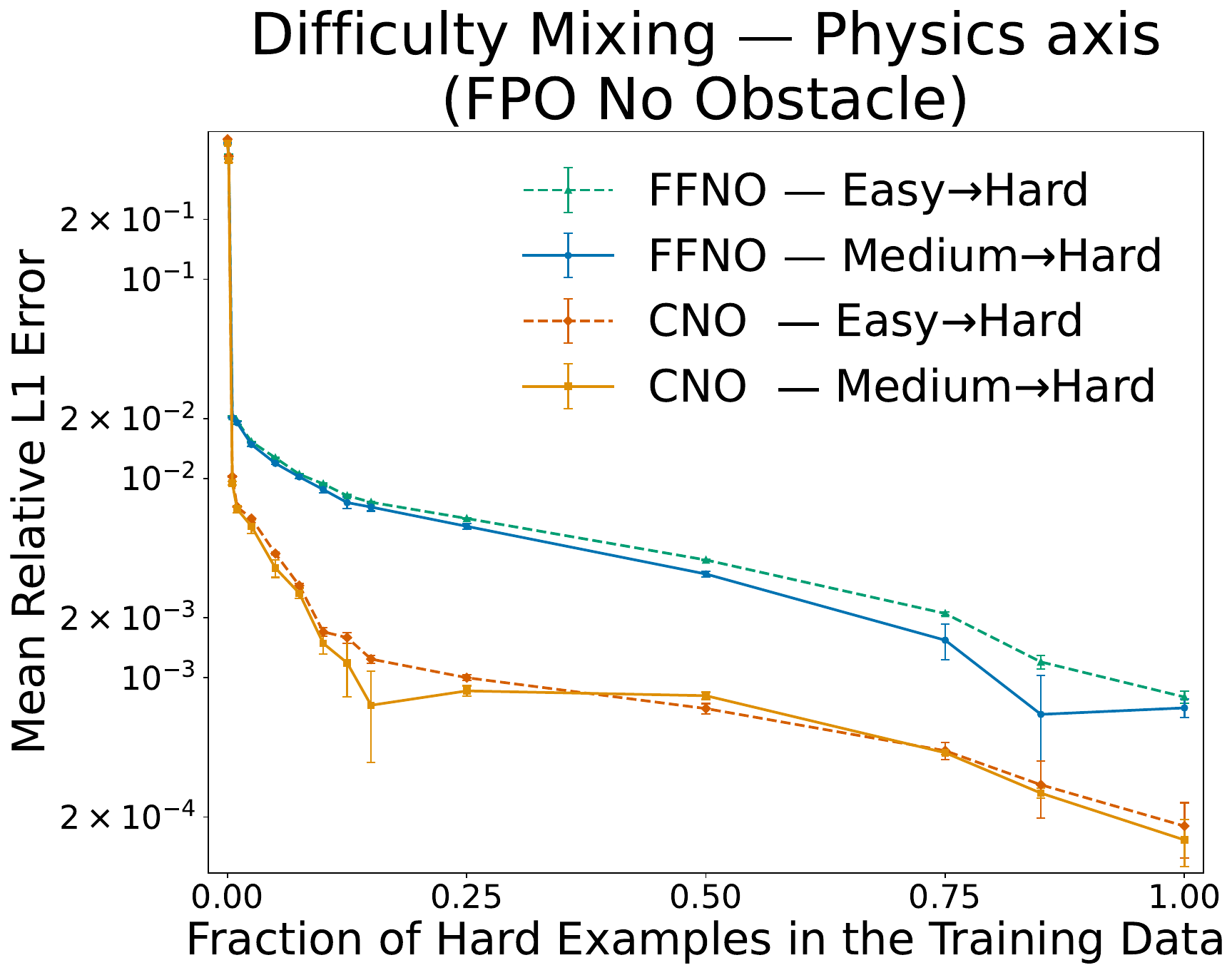}
\hfill
\includegraphics[width=0.495\linewidth]{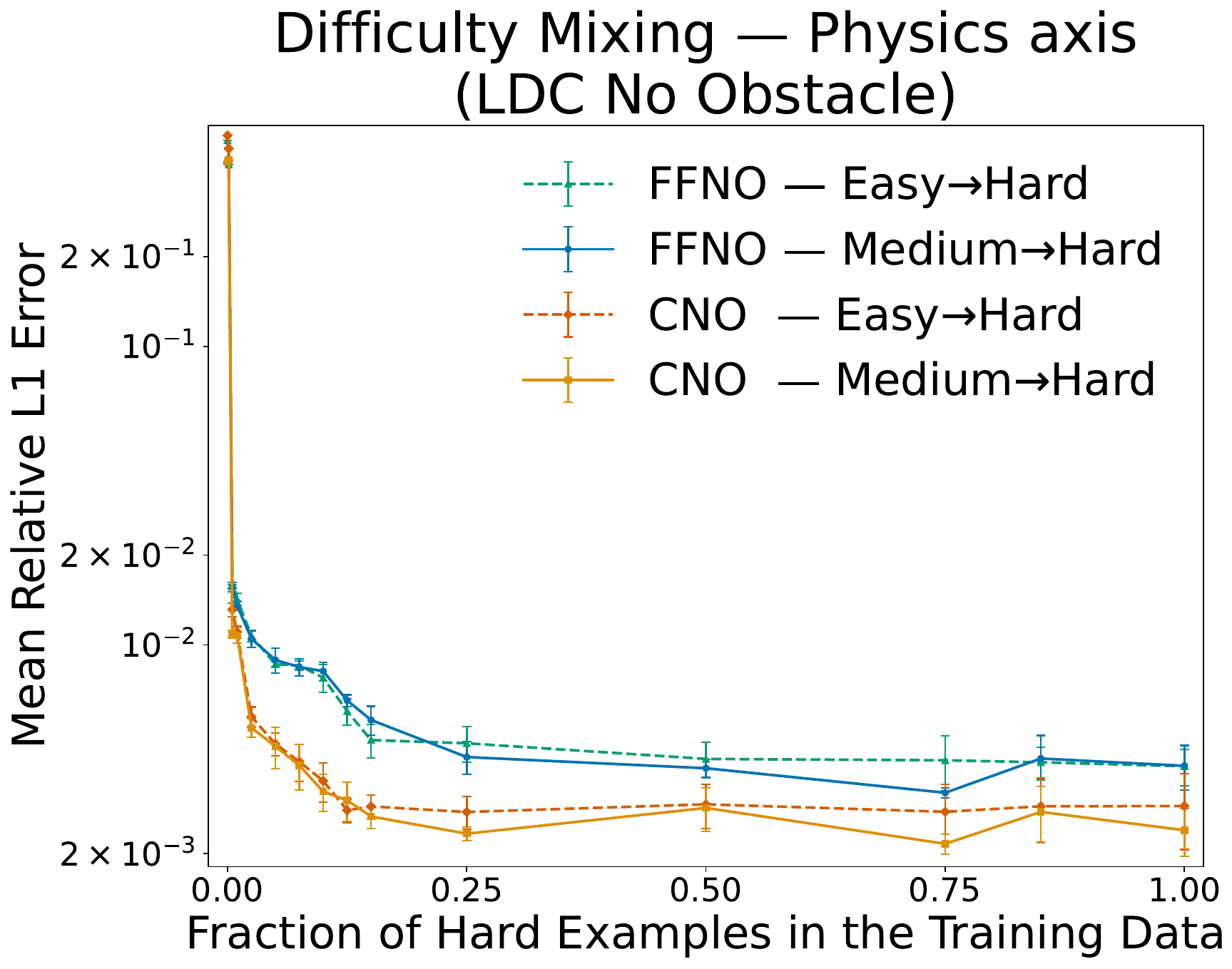}
\includegraphics[width=0.495\linewidth]{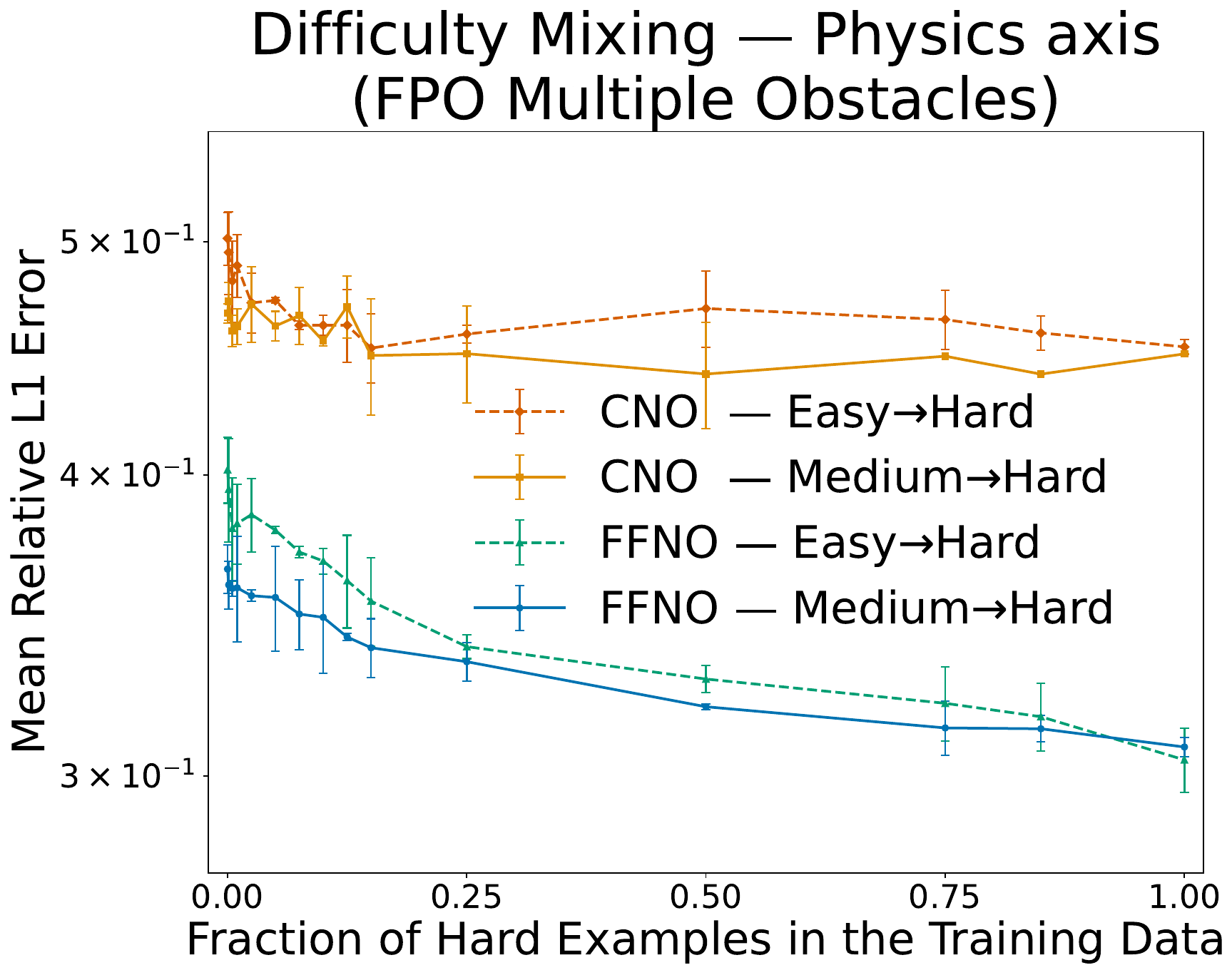}
\hfill
\includegraphics[width=0.495\linewidth]{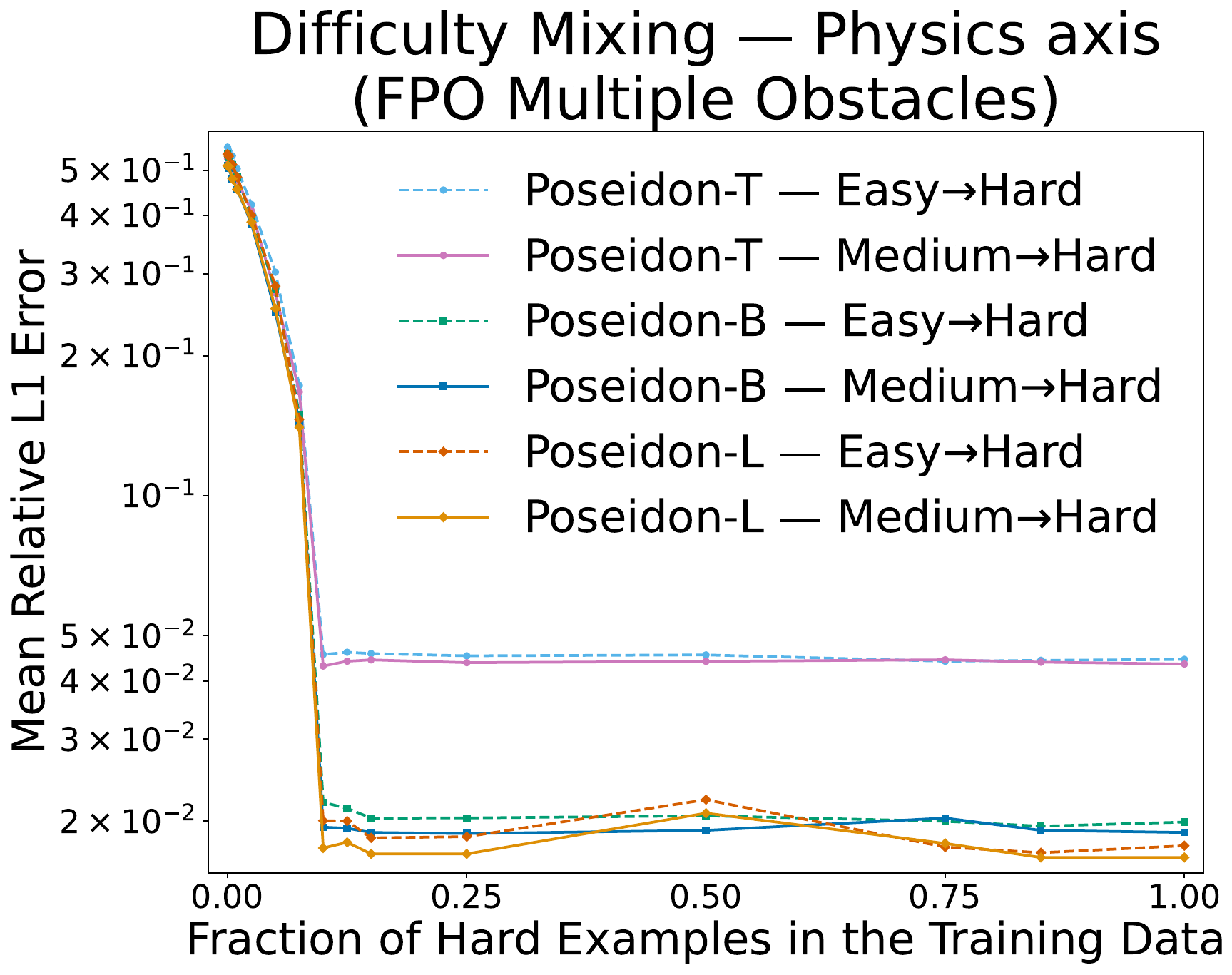}
  % \begin{subfigure}{0.495\linewidth}
  %   \centering
  %   \includegraphics[width=\linewidth]{Images/alpha_mixing_changing_physics_regular_baselines.pdf}
  %   % \caption{No Obstacle FPO (FFNO, CNO)}
  %   \label{fig:alpha-phys-regular-fpo}
  % \end{subfigure}\hfill
  % \begin{subfigure}{0.495\linewidth}
  %   \centering
  %   \includegraphics[width=\linewidth]{Images/alpha_mixing_changing_physics_regular_baselines_LDC.pdf}
  %   % \caption{No Obstacle LDC (FFNO, CNO)}
  %   \label{fig:alpha-phys-regular-ldc}
  % \end{subfigure}
  % \medskip
  % \begin{subfigure}{0.495\linewidth}
  %   \centering
  %   \includegraphics[width=\linewidth]{Images/alpha_mixing_changing_physics_baselines.pdf}
  %   % \caption{Multi-obstacle FPO (FFNO, CNO)}
  %   \label{fig:alpha-phys-mhole-fpo-cnoffno}
  % \end{subfigure}\hfill
%   \begin{subfigure}{0.495\linewidth}
%     \centering
% \includegraphics[width=\linewidth]{Images/alpha_mixing_changing_physics_poseidon.pdf}
%     % \caption{Multi-obstacle FPO (Poseidon T/B/L)}
%     \label{fig:alpha-phys-mhole-fpo-poseidon}
%   \end{subfigure}
\caption{
    {\bf Performance on hard (high Re) examples while varying the data composition.}
    We fix the total number of training examples to 800 and show the error of various models as the fraction of the data consisting of high Re ($\in[8000,10000]$) examples increases.
    Here the easy examples and medium examples are low Re ($\in[100,1000]$) and medium Re ($\in[2000,4000]$), respectively.
    The two row evaluates supervised models on no-obstacle FPO~(left) and LDC~(right), the bottom left evaluates supervised models on flows past multiple objects, and the right evaluates multiple Poseidon FMs on flows past multiple objects.
    Across all results we observe that a small fraction of lower difficulty examples is able to recover much of the performance of neural PDE solvers trained on solely hard (target) examples.
    }
% Just a small number of hard cases captures most of the gains, with medium-physics additions especially sample-efficient.}
  \label{fig:alpha-physics}
\end{figure}

We now turn to our empirical investigation, in which we evaluate several supervised and foundation models while varying the difficulty composition of their training and fine-tuning data along the difficulty axes described in Section~\ref{sec:axes}.
This results in three key takeaways:
mixing in lower difficulty data can be sufficient for strong performance~(Section~\ref{sec:simpler}), it can be beneficial to mix in a few medium difficulty examples rather than many easy ones~(Section~\ref{sec:intermediate}), and there is potential for ``foundation datasets'' that have strong few-shot performance on diverse data, as suggested via few-shot evaluations on FlowBench~(Section~\ref{sec:foundation}).
Crucially, throughout we are interested in the model's performance a {\em target distribution} consisting only of the relevant axis's hard examples, which we estimate by evaluating on a held out set.
For example, if we train on mixture of no-obstacle, single-obstacle, and multi-obstacle training examples, we report performance on a test set drawn from only the latter's distribution.

The specific supervised models we consider are the Convolutional Neural Operator~(CNO)~\citep{raonic2023convolutional} and the Factorized Fourier Neural Operator~(FFNO)~\citep{tran2021factorized}, which have demonstrated strong performance on several benchmarks~\citep{ohana2024well,tali2024flowbenchlargescalebenchmark,dauner2024resffno,takamoto2022pdebench,koehler2024apebench}.
These two models are trained from scratch on the different training mixtures we consider.
To see whether our findings continue to hold in the higher performance regimes enabled by large-scale multi-physics pretraining, we also consider the Poseidon family of FMs trained on diverse PDE families~\citep{herde2024poseidon}, evaluating three variants:
Tiny~(21M parameters), Base~(158M), and Large~(629M).
Unlike CNO and FFNO, in this case we train or {\em fine-tune} the models on our training mixure starting from a model checkpoint pretrained on diverse PDE families.
Training details of all models are reported in the Appendix.
\looseness-1

\begin{figure}[!t]
  \centering
  \includegraphics[width=0.495\linewidth]{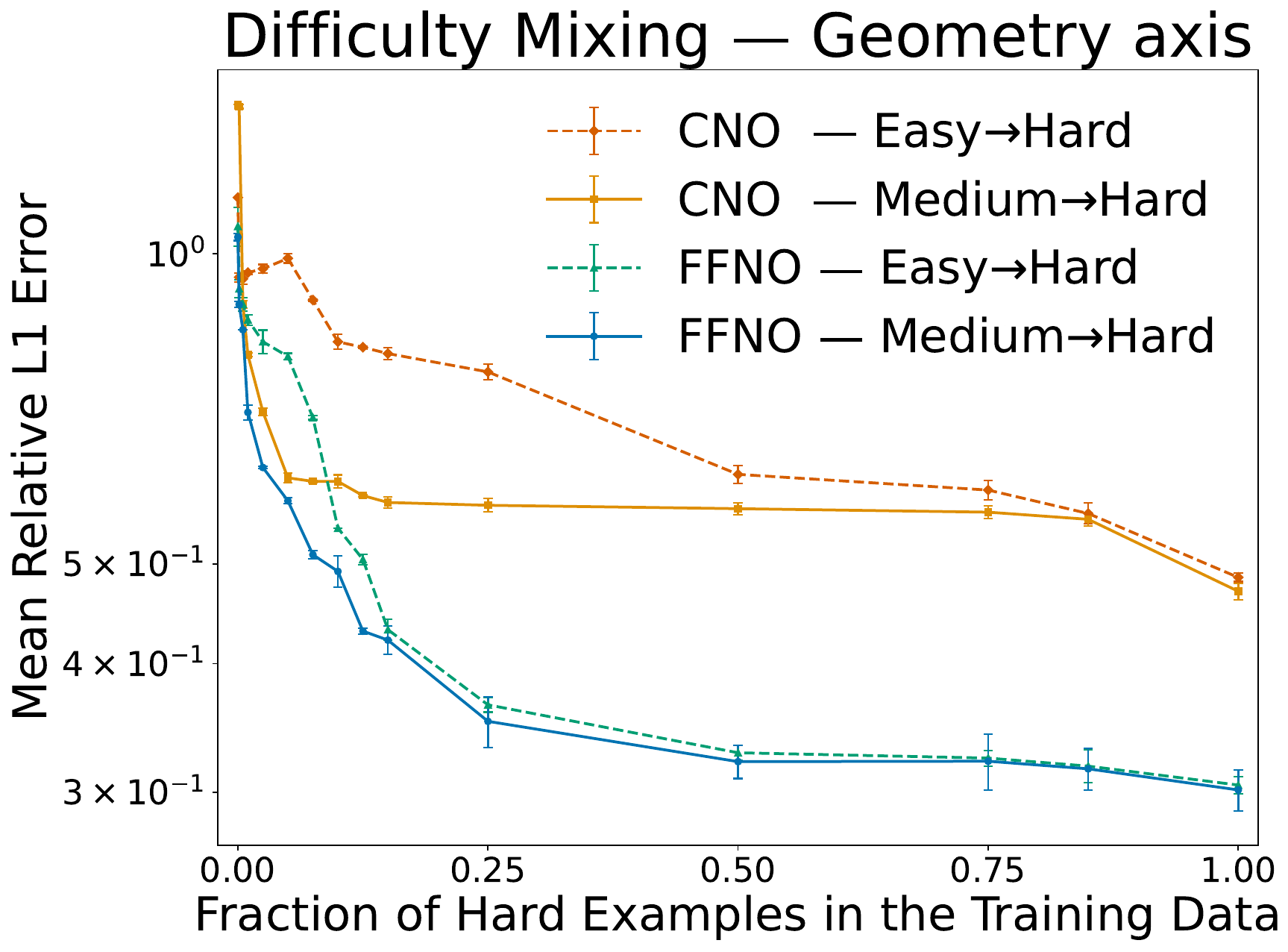}
  \hfill
\includegraphics[width=0.495\linewidth]{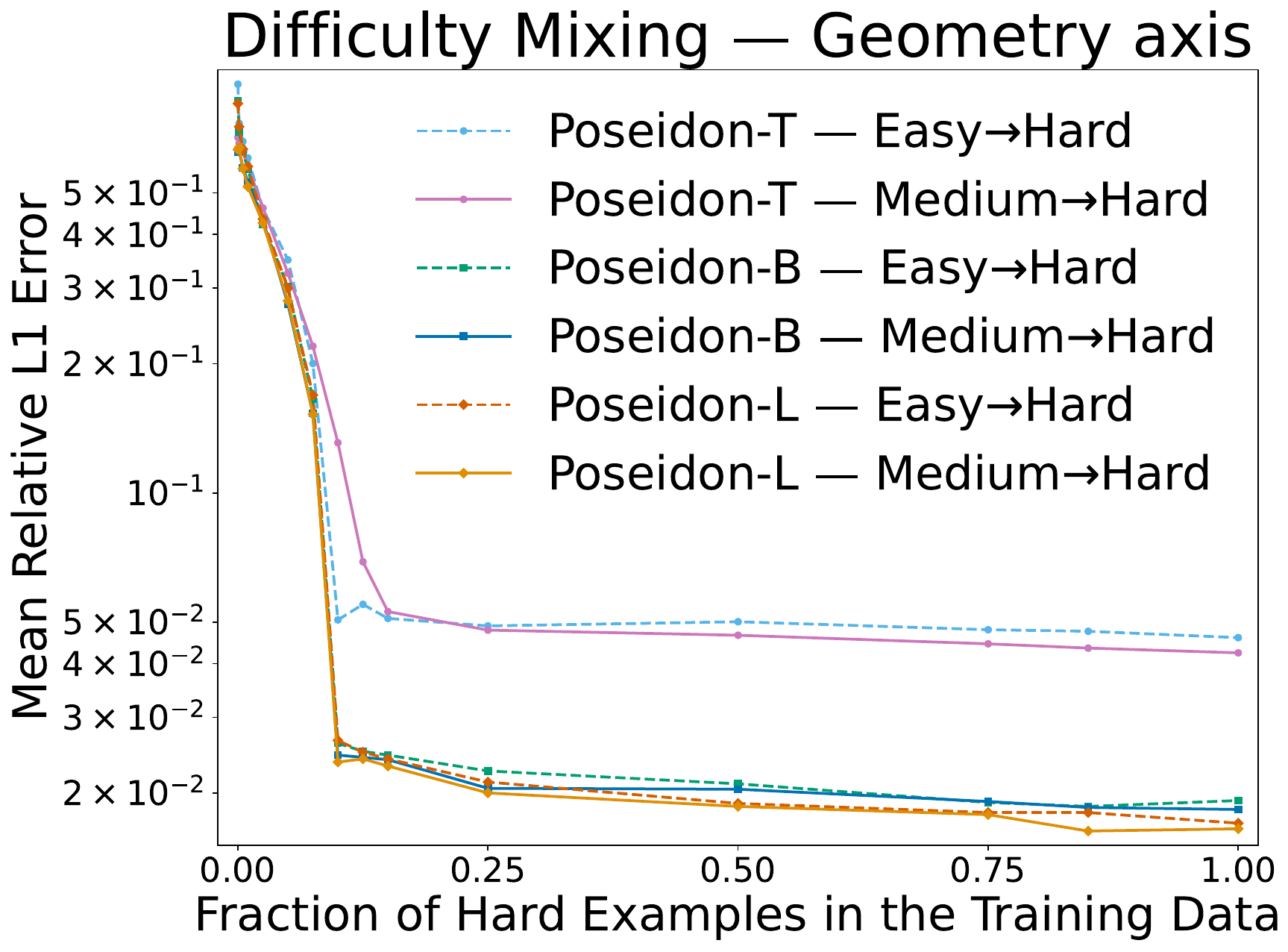}
%   \begin{subfigure}{0.495\linewidth}
%     \centering
% \includegraphics[width=\linewidth]{Images/alpha_mixing_changing_geometry_baselines.pdf}
%     % \caption{FPO (FFNO, CNO)}
%     \label{fig:alpha-geom-fpo-cnoffno}
%   \end{subfigure}\hfill
%   \begin{subfigure}{0.495\linewidth}
%     \centering
% \includegraphics[width=\linewidth]{Images/alpha_mixing_changing_geometry_poseidon.pdf}
%     % \caption{FPO (Poseidon T/B/L)}
%     \label{fig:alpha-geom-fpo-poseidon}
%   \end{subfigure}
  \caption{\textbf{Performance on hard (multi-obstacle) FPO while varying data composition.}
  The total number of training examples is fixed to 800 and we evaluate using varying fractions of zero obstacle~(easy) and single obstacle~(medium) simulations in the training data.
  As with varying Re, for both supervised models~(left) and Poseidon FMs~(right), a small number of lower difficulty examples suffices to recover most of the performance of models trained on entirely hard examples.\looseness-1
}
  \label{fig:alpha-geometry}
\end{figure}

\begin{figure}[!t]
  \centering
\includegraphics[width=0.495\linewidth]{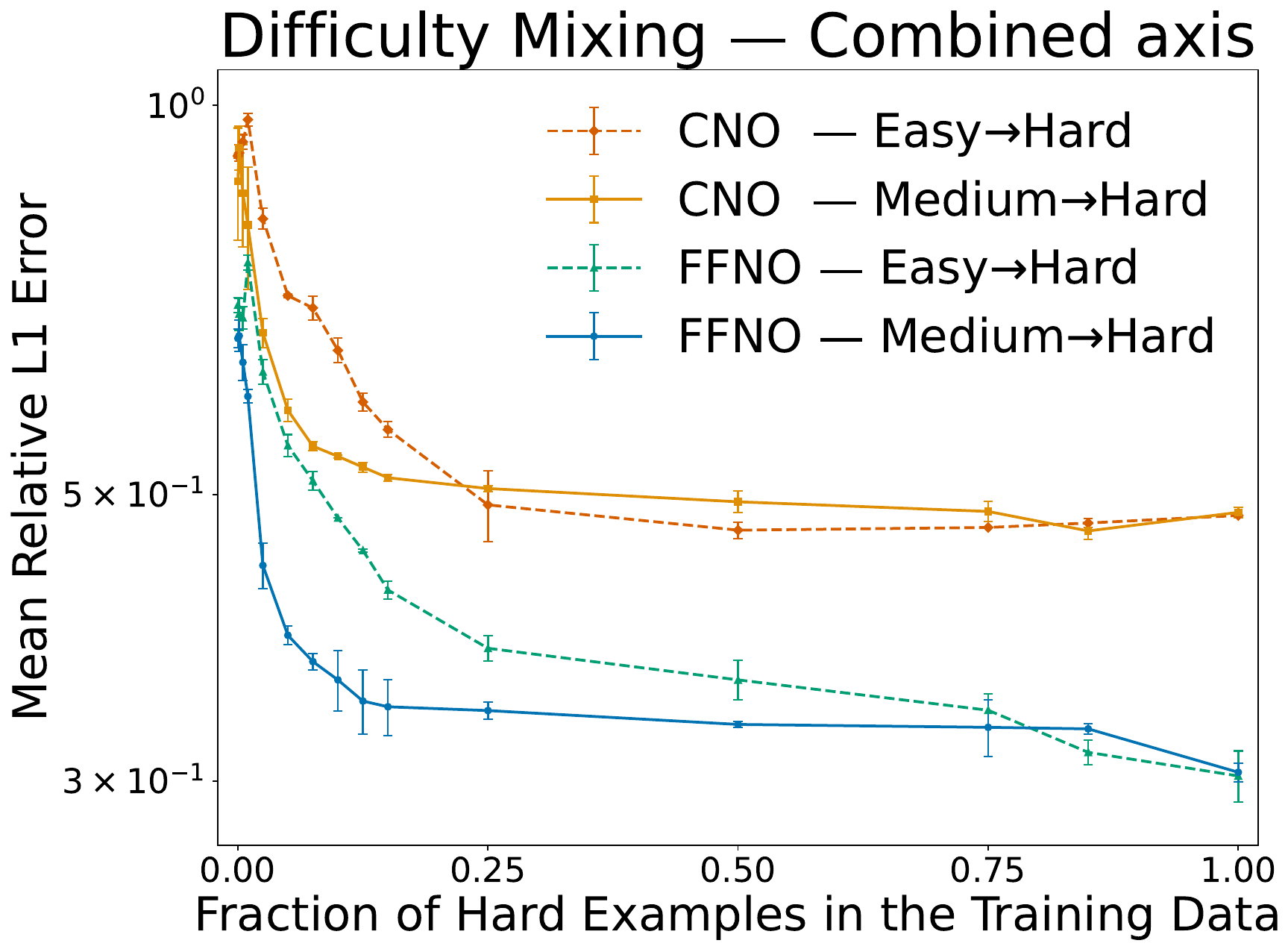}
  \hfill
  \includegraphics[width=0.495\linewidth]{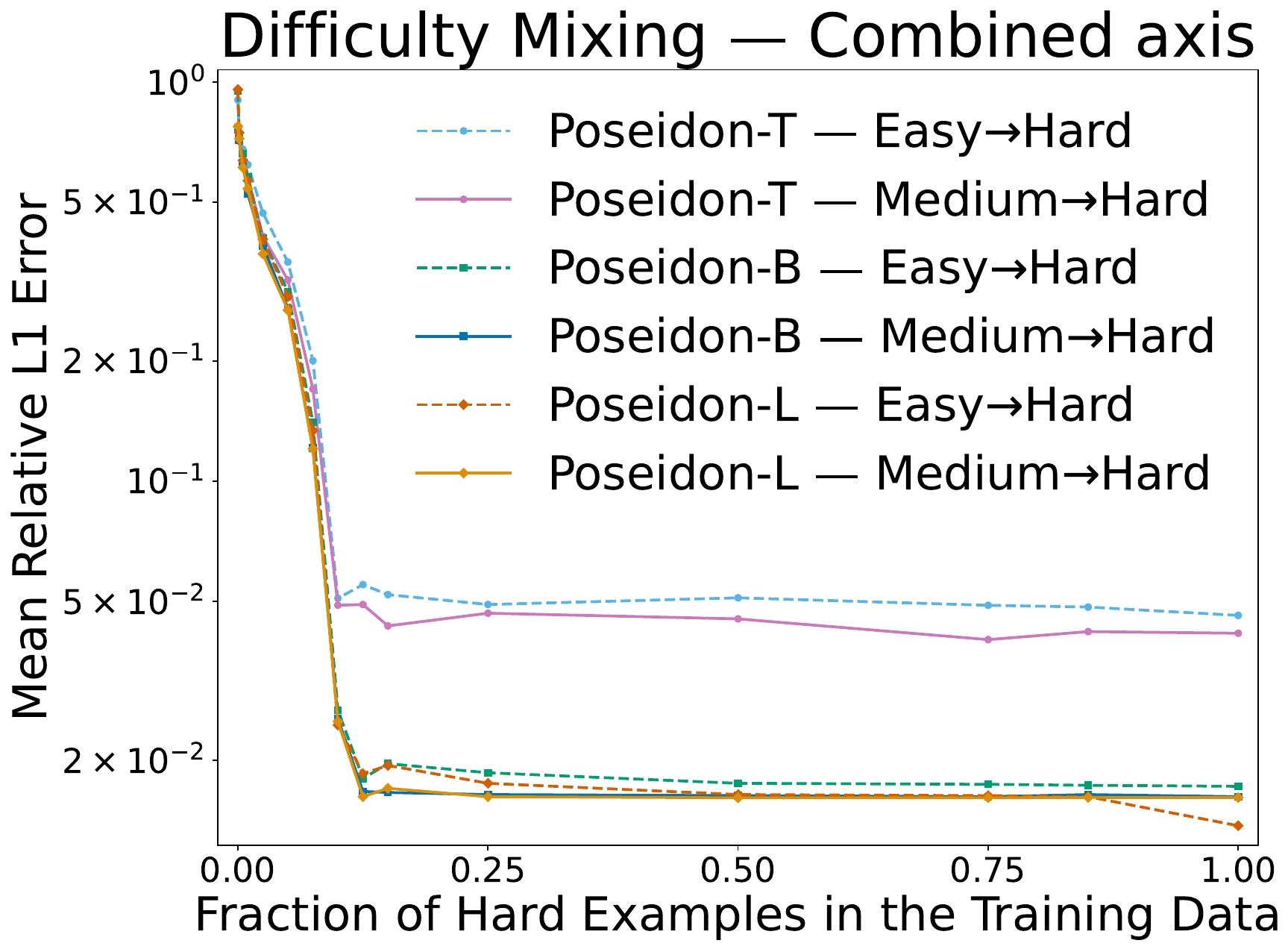}
  % \begin{subfigure}{0.495\linewidth}
  %   \centering
  %   \includegraphics[width=\linewidth]{Images/alpha_mixing_changing_physics_geometry_baselines.pdf}
  %   % \caption{FPO (FFNO, CNO)}
  %   \label{fig:alpha-physgeom-fpo-cnoffno}
  % \end{subfigure}\hfill
  % \begin{subfigure}{0.495\linewidth}
  %   \centering
  %   \includegraphics[width=\linewidth]{Images/alpha_mixing_changing_physics_geometry_poseidon_CORRECTED_v2.pdf}
  %   % \caption{FPO (Poseidon T/B/L)}
  %   \label{fig:alpha-physgeom-fpo-poseidon}
  % \end{subfigure}
  \caption{\textbf{Performance on hard (multi-obstacle and high Re) FPO while varying data composition along both physics and geometry.}
  The total number of training examples is fixed to 800 and we evaluate using varying fractions of zero obstacle low Re flow~(easy) and single obstacle medium Re flow~(medium) simulations in the training data.
  As with varying Re and geometry separately, for both supervised models~(left) and Poseidon FMs~(right), a small number of lower difficulty examples suffices to recover most of the performance of models trained on entirely hard examples.\looseness-1}
  \label{fig:alpha-physgeom}
\end{figure}

\vspace{-2mm}
\subsection{Training on simpler examples goes a long way}\label{sec:simpler}
\vspace{-1mm}

We start with {\bf difficulty-mixing} evaluations, in which we fix the {\em total} number of training points to $n=800$ and vary the proportion allocated to hard examples from the target distribution.
Our main finding is that adding a small set of hard examples to otherwise lower difficulty~(easy and medium) training data is sufficient to recover most of the performance of training on a dataset where all 800 examples are hard.
Below we discuss how this manifests along specific difficulty axes.
Note that the total number $n=800$ of training points was determined by training only on hard data using several candidate budgets $n$ and finding that the test error plateaued after around 800 examples;
we standardize this budget throughout this subsection, although as discussed in Figure~\ref{fig:poseidon-budget} our main finding holds for other budgets as well.\looseness-1

\begin{enumerate}[leftmargin=*,topsep=0mm,itemsep=1pt,parsep=0pt]
    \item {\bf Physics axis} (Fig.~\ref{fig:alpha-physics}): 
    While models trained on lower difficulty examples do poorly on the hard~(high Re) test examples, replacing just 10\% of them by target distributions examples recovers most of the benefit of training fully on the latter.
    Notably, using the numbers in Figure~\ref{fig:simtime-fpo} we see that the former involves 8.9$\times$ less compute time.
    At 10\% hard examples, Poseidon-B typically reduces error by about $96\%$ at 10\% hard data, while CNO and FFNO show $\approx 98\%$ reductions in the no-obstacle and $\approx 6\%$ in the multi-obstacle setting.
    Increasing the proportion of hard examples provides incremental gains until 25\% and plateaus after.\looseness-1
    \item {\bf Geometry axis} (Fig.~\ref{fig:alpha-geometry}): The same pattern when composing multi-object FPO with flows past zero or one objects: 
    the main improvement for CNO, FFNO, and Poseidon is obtained when only 10\% of the data is from the target distribution.
    In particular, at that percentage Poseidon-B improves  by roughly $96$–$97\%$ in terms of error relative to training on all-easy examples.
    Additional hard data yields only modest improvement.
    \item {\bf Combined axis} (Fig.~\ref{fig:alpha-physgeom}):
    We observe similar behavior when varying along the combined domain geometry and problem physics axis, with most of the benefit of training on the target distribution obtained with 10\% examples and improving only modestly afterwards.
\end{enumerate}

In summary, across all three difficulty axes and all model families,
we consistently find that a small hard fraction (often around 10\%) is enough to obtain most of the performance of hard-only training; 
results change only marginally
beyond $\approx$25\% data from the target distribution.\looseness-1
% \noindent Early gains dominate and medium data are more sample-efficient than easy-only mixes; next we fix $N_{\text{hard}}{=}200$ and scale the easy/medium pool (Section~\ref{sec:intermediate}).
% It is also worth noting that---here and elsewhere---both the CNO and FFNO models consistently do worse than Poseidon.
% This could be attributed to the fact that both models are trained from scratch, whereas the Poseidon model 
% is pretrained on multiple physics datasets and perhaps benefits from such pre-training. 
% This shows that for such pre-trained models, we may be more likely to get away with pre-generating and training with simpler but easier to generate examples.

\vspace{-2mm}
\subsection{Cost-effectiveness of pre-generating fewer medium difficulty examples}\label{sec:intermediate}
\vspace{-1mm}

% -----------------------------
% Scaling — Changing Physics
% -----------------------------
\begin{figure}[!t]
  \centering
  \begin{subfigure}{.495\linewidth}
    \centering
    \includegraphics[width=\linewidth]{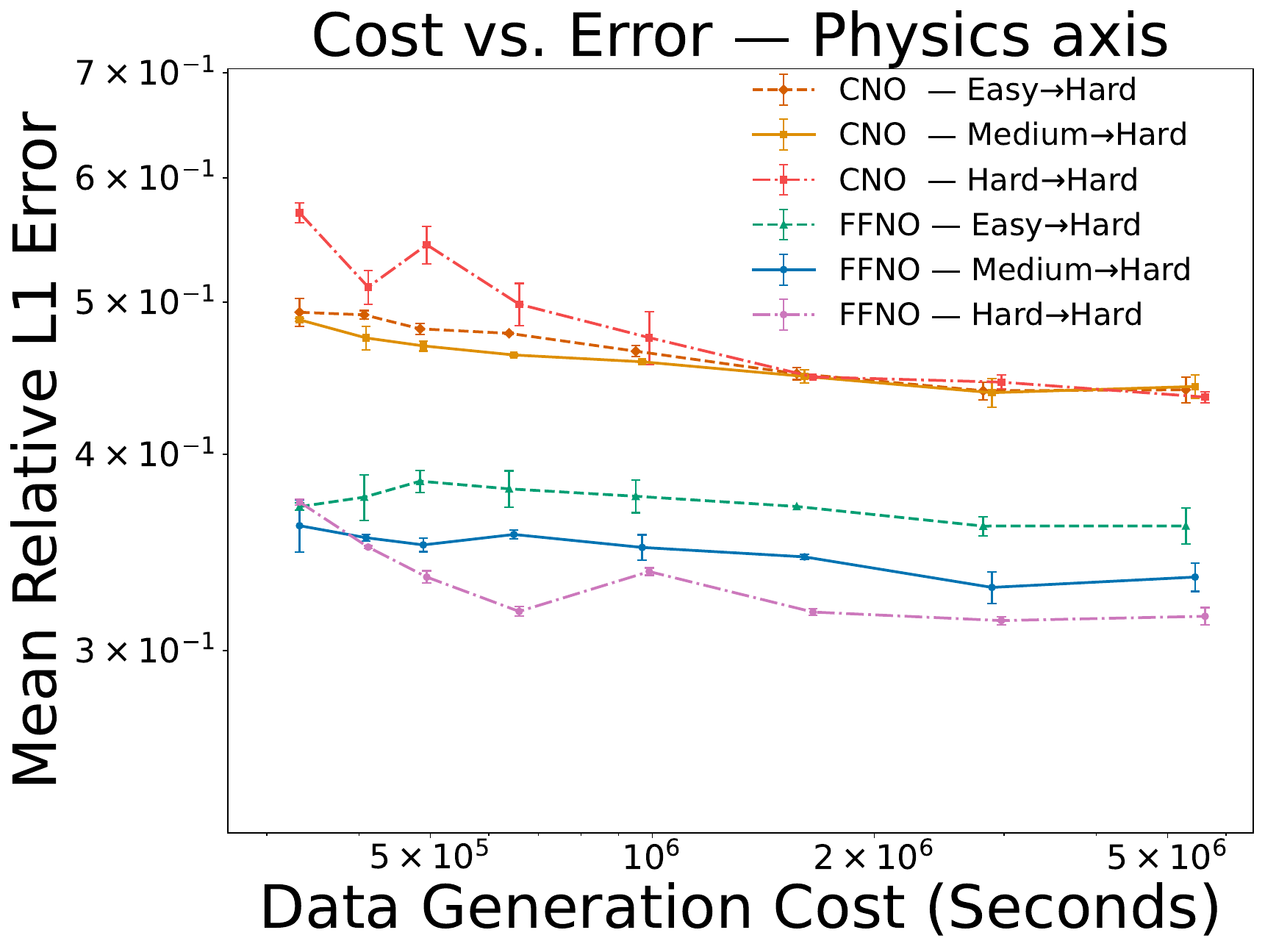}
    % \caption{FPO (FFNO, CNO)}
  \end{subfigure}\hfill
  \begin{subfigure}{.495\linewidth}
    \centering
    \includegraphics[width=\linewidth]{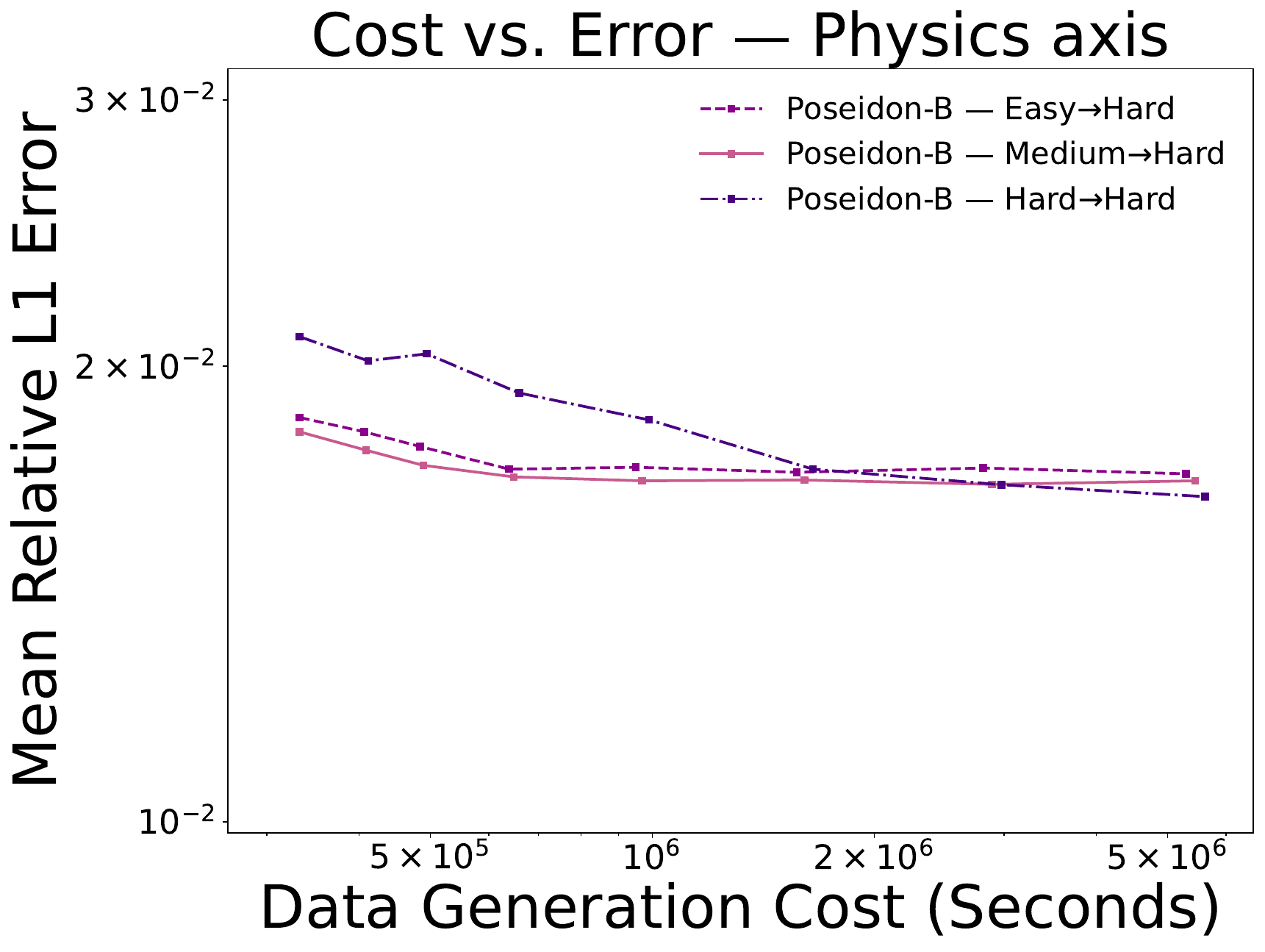}
    % \caption{FPO (Poseidon-B)}
  \end{subfigure}
  \caption{
    {\bf Comparing data generation cost vs. error while augmenting hard~(multi-obstacle high Re) FPO examples} with easy~(multi-obstacle low Re), medium~(multi-obstacle medium Re), and hard~(multi-obstacle high Re) examples.
    We fix the number of hard examples to 200 and plot the compute required to generate them and between 1 and 3200 lower and equal difficulty examples.
    For both supervised models~(left) forand Poseidon-B~(right), generating medium difficulty data has a generally more favorable tradeoff, achieving the same or lower error at the same budget. At sufficiently large compute budgets, however, training exclusively on hard data (hard to hard) yields the lowest error.}
  \label{fig:scaling-phys}
\end{figure}

Having demonstrated that low-cost simulation data can be added to just a few (harder-to-obtain) examples from the target distribution to recover much of the performance trained solely on the hard examples, we now examine the cost vs. error tradeoffs of using data at different points on the difficulty axis.
In particular, we examine whether there are regimes in which it is favorable to generate and train on medium difficulty (e.g. single-obstacle or intermediate Re) examples rather than easy examples.
To do so we fix the number of hard examples to 200 and vary the number of lower difficulty examples added to the training mix between 1 and 3200. 
For completeness, we also evaluate a \emph{hard-on-hard} variant that augments the $N_{\text{hard}}{=}200$ seed with additional target-distribution (high-Re, multi-obstacle) samples. This setting delivers the lowest error per added sample but at the highest pre-generation cost, so we use it primarily as an upper-bound reference when comparing cost-normalized tradeoffs to medium- and easy-on-hard mixes.

Since medium difficulty examples are more costly to generate than easy ones, we study how the error varies as a function of the pre-generation cost.
Our main finding is that there indeed are many pre-generation budgets at which the error obtained by training on medium difficulty examples is lower than that obtained by training on (more) easy examples.
Below we discuss the extent to which this holds along specific difficulty axes.\looseness-1

% We study \textbf{medium-vs-easy scaling}: we fix the \emph{hard} seed at $N_{\text{hard}}=200$ and grow an auxiliary pool from either \emph{medium} or \emph{easy} difficulty, always evaluating on the hard test distribution (nMAE, fraction units).

% \textbf{Main result.} Across axes and model families, adding \emph{medium} examples delivers the early, dominant gains. With only $\sim$200 auxiliary samples, FFNO/CNO typically realize single- to mid–double-digit error reductions over easy, while Poseidon-B gains are smaller but consistent; the first few hundred medium examples usually capture $\approx$80--90\% of the eventual improvement.

\begin{figure}[!t]
  \centering
  \begin{subfigure}{.495\linewidth}
    \centering
    \includegraphics[width=\linewidth]{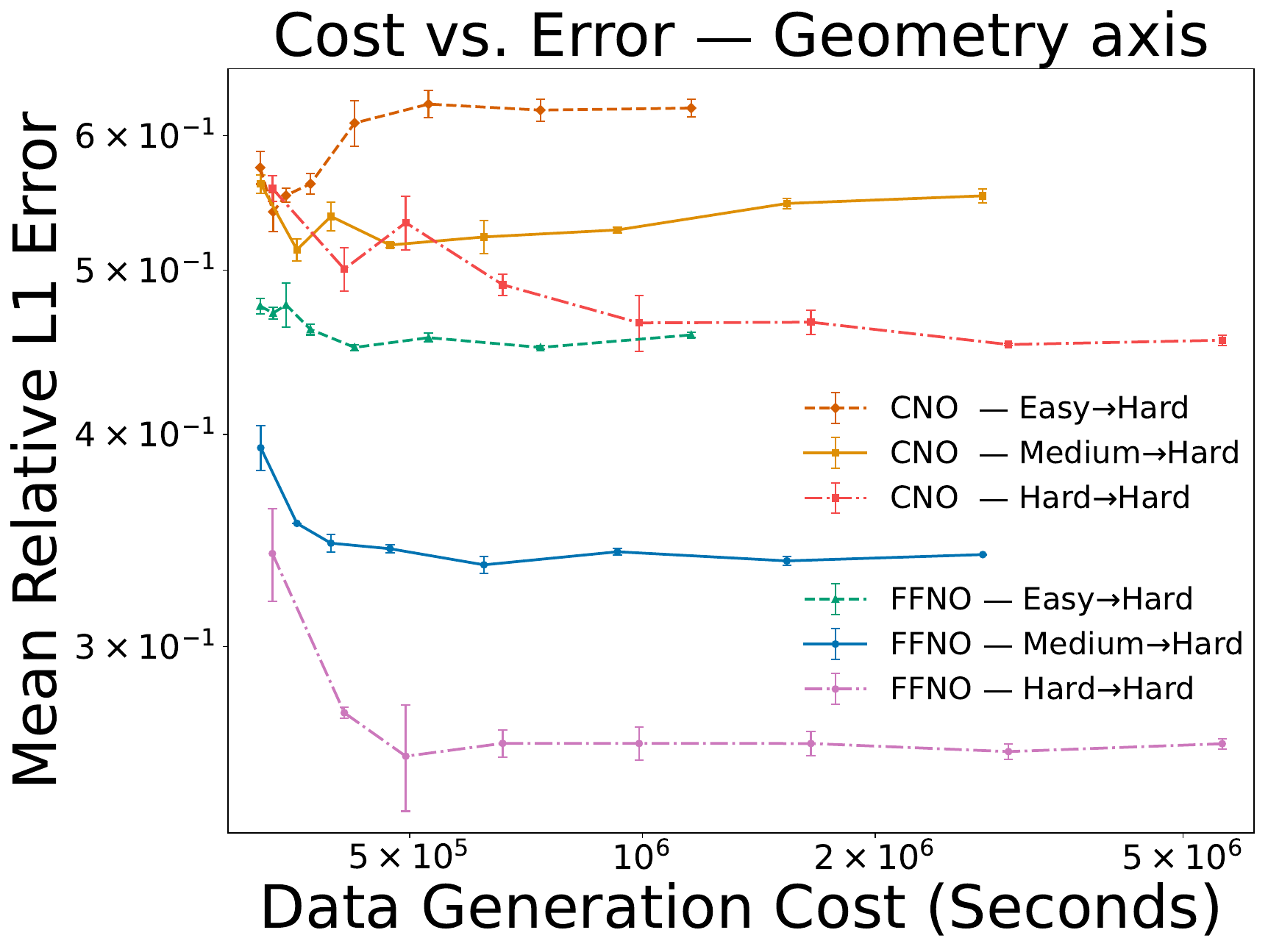}
    % \caption{FPO (FFNO, CNO)}
  \end{subfigure}\hfill
  \begin{subfigure}{.495\linewidth}
    \centering
    \includegraphics[width=\linewidth]{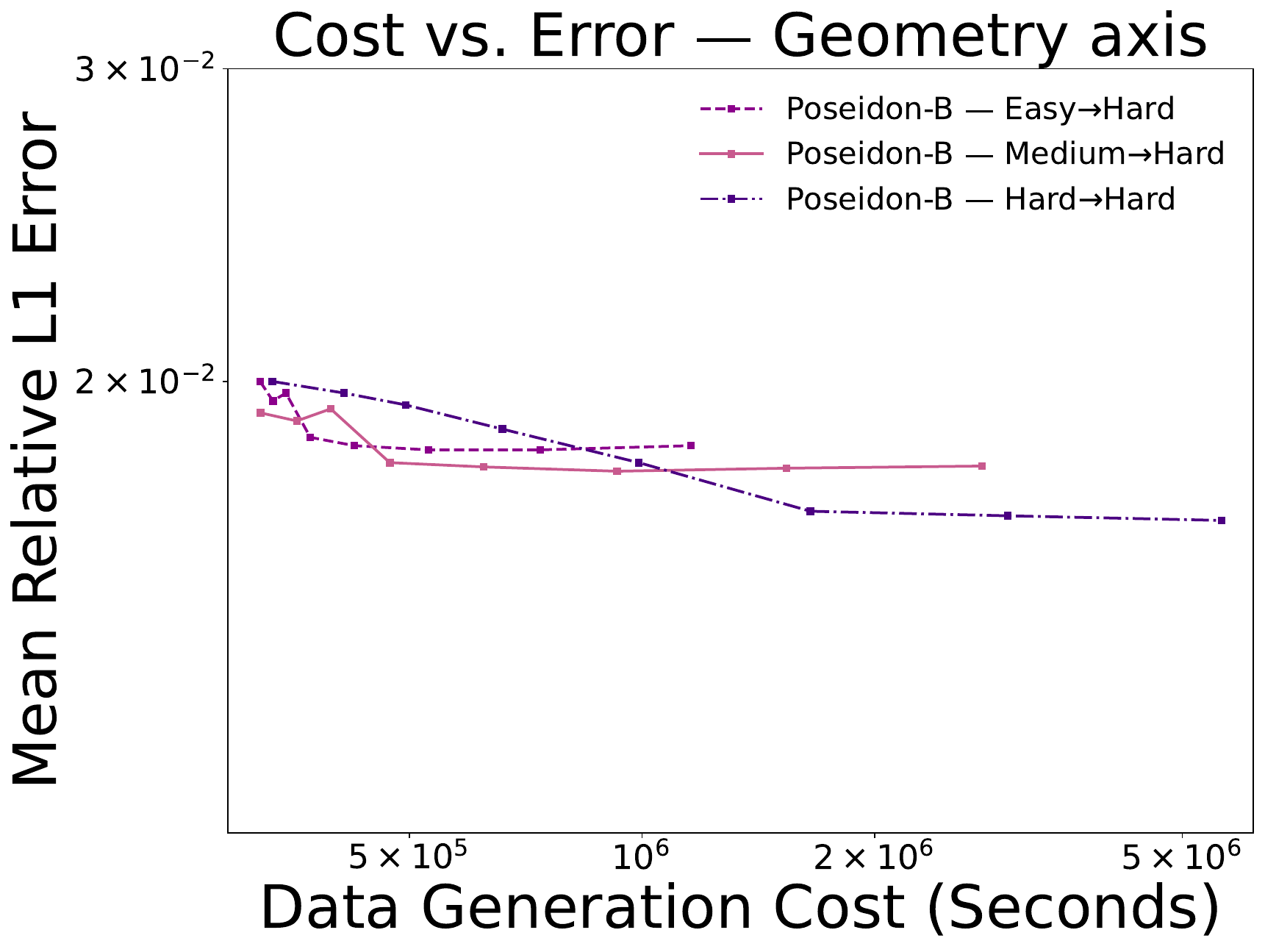}
    % \caption{FPO (Poseidon-B)}
  \end{subfigure}
  \caption{{\bf Comparing data generation cost vs. error while augmenting hard~(multi-obstacle) FPO} with easy~(no obstacle), medium~(single obstacle), and hard~(multi-obstacle) examples.
    We fix the number of hard examples  to 200 and plot the compute required to generate them and between 1 and 3200 lower and equal difficulty examples.
    For supervised models, generating medium difficulty data has a generally more favorable tradeoff, achieving the same or lower error at the same budget;
    for Poseidon-B, generating medium data is more cost-effective given 5e5 seconds or more time for pre-generation. At sufficiently large compute budgets, however, training exclusively on hard data (hard to hard) yields the lowest error.\looseness-1
    }
  \label{fig:scaling-geom}
\end{figure}

\begin{figure}[!t]
  \centering
  \begin{subfigure}{.47\linewidth}
    \centering
    \includegraphics[width=\linewidth]{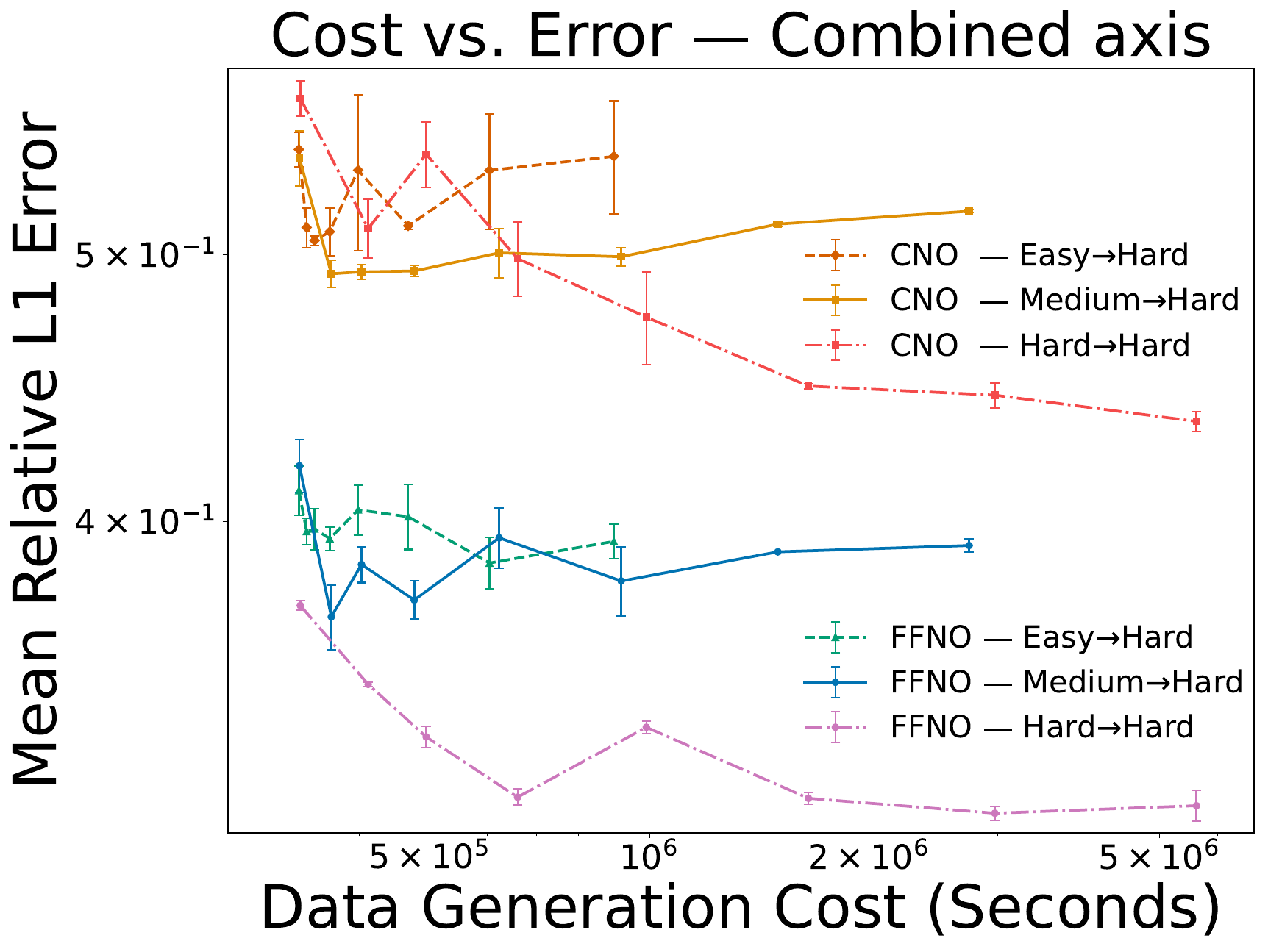}
    % \caption{FPO (FFNO, CNO)}
  \end{subfigure}\hfill
  \begin{subfigure}{.47\linewidth}
    \centering
    \includegraphics[width=\linewidth]{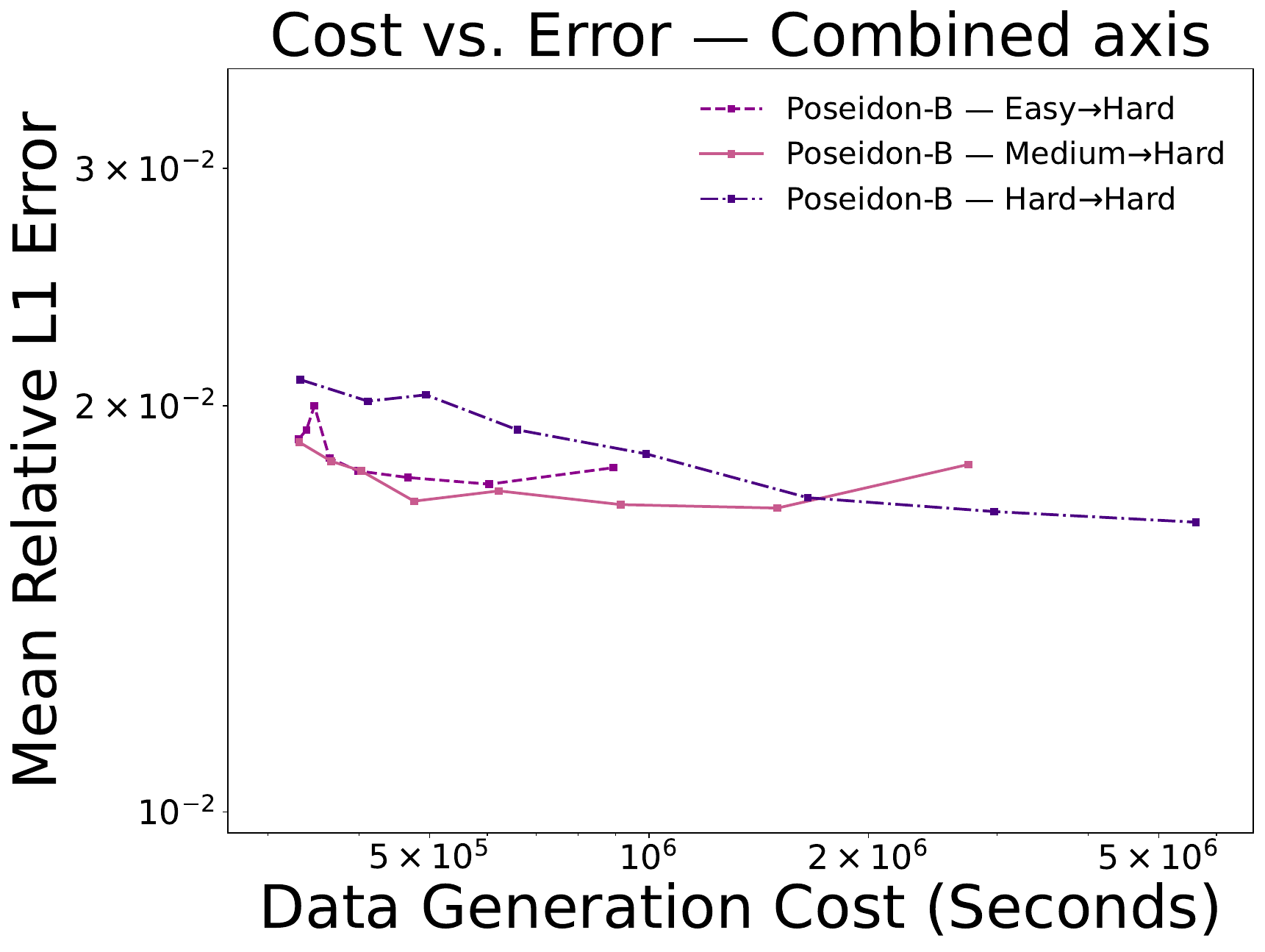}
    % \caption{FPO (Poseidon-B)}
  \end{subfigure}
  \caption{{\bf Comparing data generation cost vs. error while augmenting hard~(multi-obstacle high Re) FPO examples} with easy~(no obstacle low Re), medium~(single obstacle medium Re), and hard~(multi-obstacle high Re) examples.
    We fix the number of hard examples to 200 and plot the compute required to generate them and between 1 and 3200 lower and equal difficulty examples.
    For both supervised models and Poseidon-B, generating medium difficulty data has a generally more favorable tradeoff, achieving the same or lower error at the same budget as easy data.
    However, in all cases, too many lower difficulty examples can reduce performance. At sufficiently large compute budgets, training exclusively on hard data (hard to hard) yields the lowest error.
    }
  \label{fig:scaling-both}
  % \vspace{-5mm}
\end{figure}

\begin{figure}[!htbp]
  \centering

  % Left column
  \begin{minipage}[t]{0.47\linewidth}
    \centering
    \includegraphics[width=\linewidth]{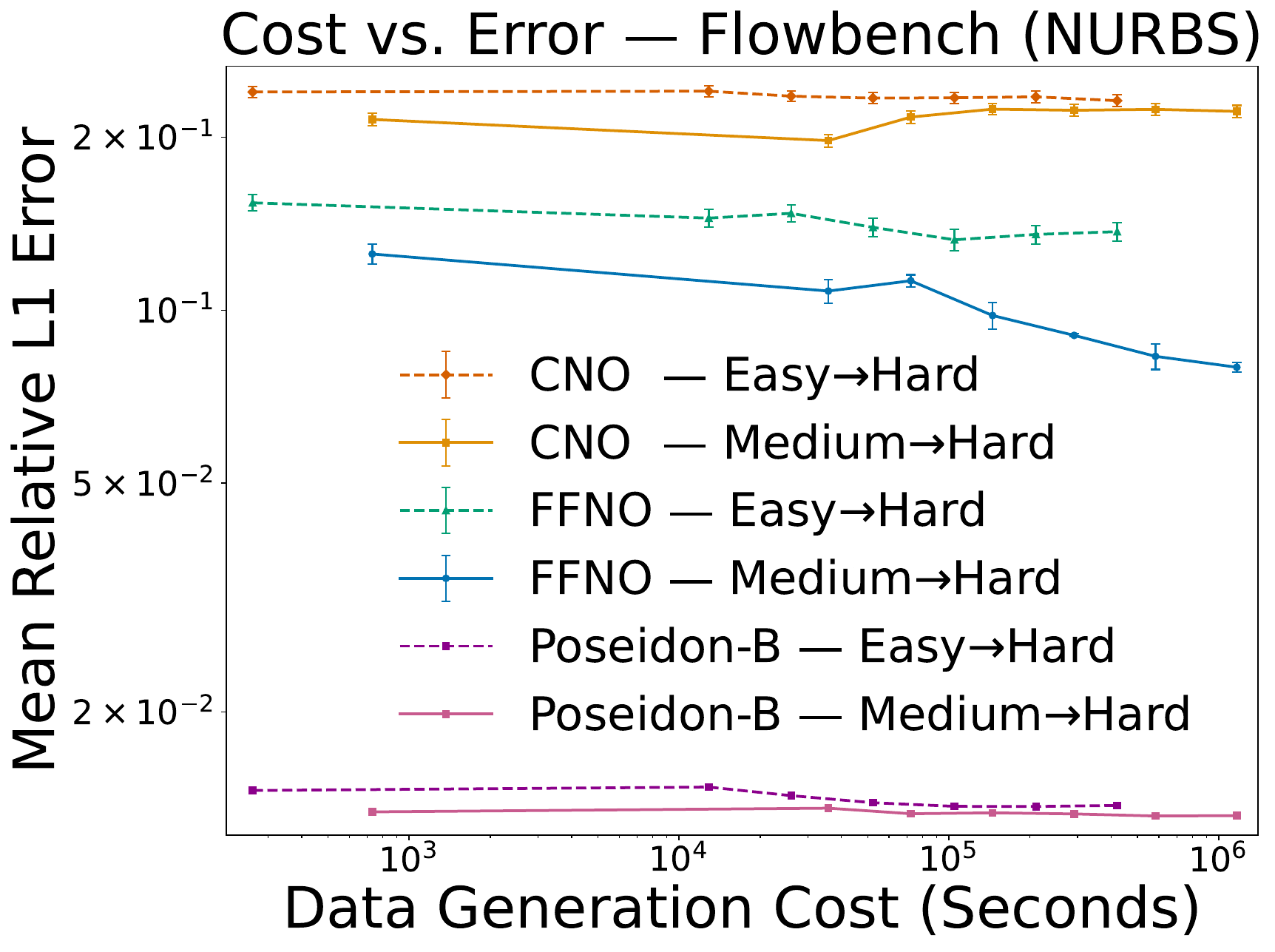}
    \caption{
        {\bf Performance on FlowBench's NURBS FPO simulations} when 200 target examples are augmented with 1-3200 zero obstacle FPO~(easy) or single square-obstacle FPO~(medium) simulations.
        In multiple cases such, e.g. adding medium examples when training FFNO, doing this data augmentation substantially improves performance on the target FlowBench distribution.}
    \label{fig:nurbs-flowbench}
  \end{minipage}\hfill
  % Right column
  \begin{minipage}[t]{0.47\linewidth}
    \centering
    \includegraphics[width=\linewidth]{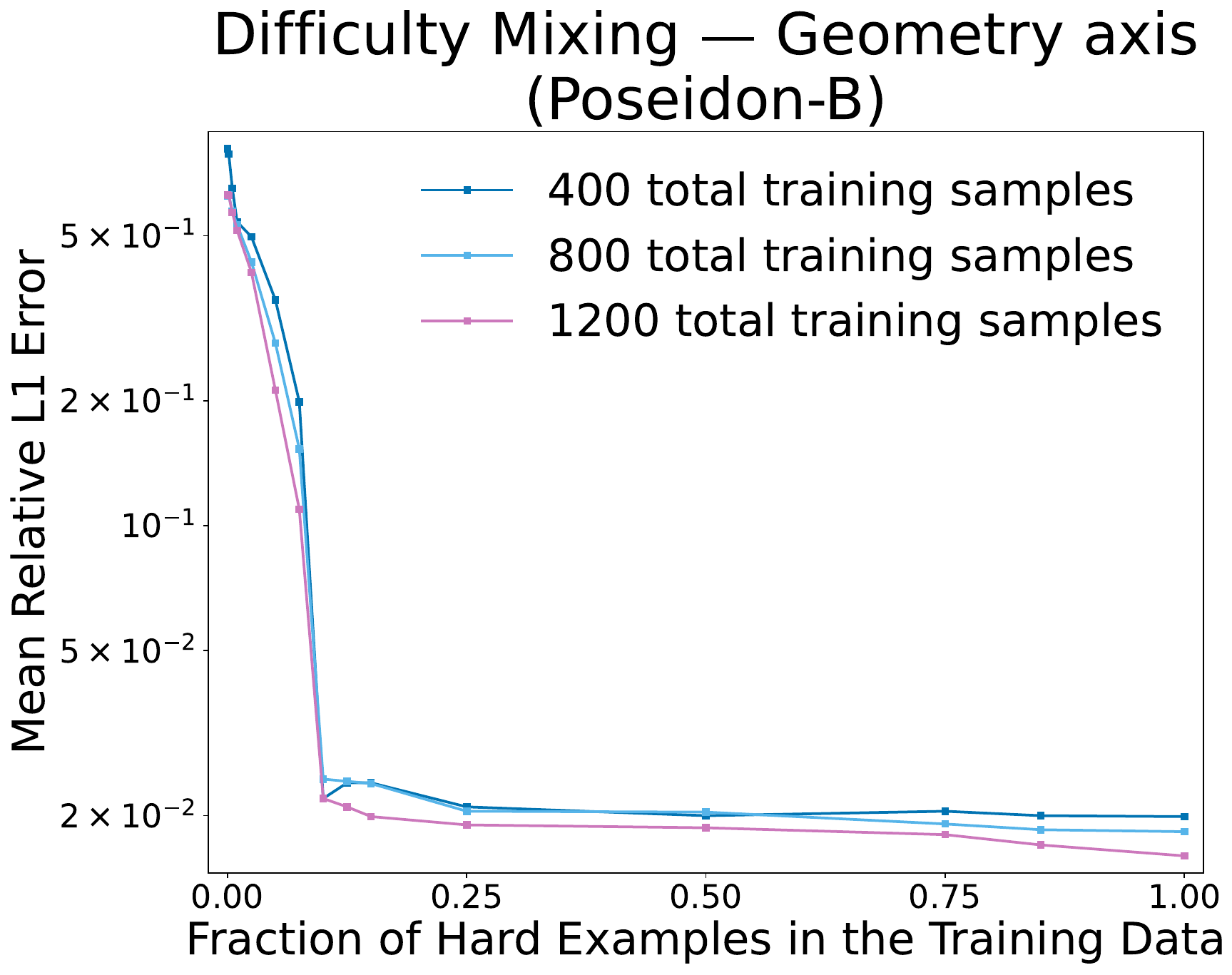}
    \caption{{\bf Performance of Poseidon-B as the fraction of target distribution data increases}.
    Each curve fixes the training set size and varies the number of medium (one obstacle) vs. hard (multi-obstacle) examples in it.
    In all three cases, most of the improvement over training on just single-obstacle examples is obtained by replacing just 10\% of the data with target distribution examples.
    }
    \label{fig:poseidon-budget}
  \end{minipage}
\end{figure}

\begin{enumerate}[leftmargin=*,topsep=0mm,itemsep=1pt,parsep=0pt]
\item \textbf{Physics axis} (Fig.~\ref{fig:scaling-phys}):
While the increase in data generation cost is small when going from low to medium Re, for both FFNO and Poseidon the error of the model trained on the latter is lower at all data generation costs evaluated, demonstrating the value of using intermediate rather than easy examples when targeting a hard distribution.
\item \textbf{Geometry axis} (Fig.~\ref{fig:scaling-geom}):
Unlike changing physics, changing the domain geometry by adding obstacles significantly increases computational cost.
Nevertheless, for the supervised models (CNO and FFNO) it is usually favorable to train on FPO simulations with one~(medium) rather than no~(easy) obstacles at all computational budgets.
For the better-performing Poseidon FM, training on medium difficulty examples is cost-effective at data-generation budget of 5e5 seconds and higher.
\item \textbf{Combined axis} (Fig.~\ref{fig:scaling-both}): when the hard examples involve a high Re flow past multiple objects, we find that augmenting with medium difficulty examples performs better than or the same as using (more)~easy examples at the same generation budget.
However, we also find that for both supervised models and the Poseidon FM that adding increasingly more low difficulty examples starts to increase the error on the target distribution, demonstrating that care needs to be taken when doing this data composition.
\end{enumerate}

In summary, across all difficulty axes we find that it is cost-effective or at least not significantly harmful to train on medium difficulty rather than easy examples, despite the former's greater generation cost.
This result demonstrates the importance of considering multiple scales of difficulty when pre-generating a data mixture for a specific high difficulty target distribution.\looseness-1

% ---------------------------------------------
% Scaling — Changing Physics \& Geometry
% ---------------------------------------------

\vspace{-2mm}
\subsection{Towards foundation datasets}\label{sec:foundation}
\vspace{-1mm}

In our last evaluation, we study the implications of multi-difficulty training for large-scale pre-generation of datasets for solving diverse PDE tasks.
As an example of the latter, we consider examples from the FlowBench dataset~\citep{tali2024flowbenchlargescalebenchmark} of flows past irregular NURBS objects~(cf. Section~\ref{sec:axes}).
Using flows past zero objects and flows past one square object as the easy and medium distributions, respectively, we show in Figure~\ref{fig:nurbs-flowbench} that adding these simpler examples can reduce the error.
This is especially pronounced in the case of FFNO when its training data is augmented with single object FPO examples.
This suggests the potential utility of pre-generating large medium difficulty datasets and reusing them on multiple other datasets, thus amortizing the pre-generation costs.
This pipeline is analogous to that of pre-training a foundation model, the cost of which is amortized as it is applied to multiple downstream tasks.\looseness-1

% Lastly, our experiments along different but interdependent axes of difficulty also reveal implications for large-scale pre-generation of datasets for PDE solvers.
% In particular, the results demonstrate that a single pre-generated dataset---e.g. single-obstacle FPO---can be used to improve performance on multiple higher difficulty settings of interest.
% For example, our single-obstacle FPO dataset can be combined with 200 FlowBench G1 NURBS examples to reduce the error of Poseidon on the latter distribution.
% Thus, while pre-generating the single-obstacle FPO dataset is expensive, its reuse via mixing with {\em multiple} other datasets is greatly multiplies of its potential downstream usefulness, in a similar way to how pre-training a foundation model is useful because it can be fine-tuned on multiple other tasks.

% For FlowBench (NURBS), we fixe $200$ examples and varied the number of easy/medium samples drawn from our pre-generated (no-obstacle and single-obstacle) FPO data. 
% Specifically, Harmonics used \emph{100} hard examples (the dataset is small once validation/testing are reserved), and NURBS used \emph{200} hard examples. 
% Across all model families, adding easy/medium data consistently reduces test error with diminishing returns, demonstrating that a single pre-generated dataset can be reused to improve performance on separate FlowBench targets as well—mirroring our main findings on FPO.

This idea of a pre-generated ``foundation dataset'' can also be used to describe much larger-scale efforts like The Well~\citep{ohana2024well}.
However, what our study demonstrates is that, just like the length and quality of web data used to train large language models matters, so does the difficulty of data pre-generated for training PDEs.
In particular, as discussed in the last section, medium~(across any axis) data can be much more effective as a mixing dataset than easy~(across any axis) data for multi-obstacle performance.
The current section further shows this for FlowBench NURBS data.
Thus, when pre-generating such large datasets, it will be important to incorporate settings that more closely approach the types of more difficult problems that will be of actual interest to future users of neural PDE solvers.\looseness-1

% !TEX root = main.tex

%\vspace{-3mm}
\section{Conclusion}
%\vspace{-2mm}

% {\color{red}
% % Discussion of improvements due to difficulty transfer in various settings, effects of different models, confirming that it's not a model capacity issue, discussion of computational costs savings and implications for pre-generation workflows, observation of generalization to both multiple objects and to FlowBench and the associated idea of ``foundation datasets'', future work
% }

This paper presents a data-centric study on the role of training data 
composition for neural PDE solvers.
Our study considers three difficulty axes comprising the {\em domain geometry} in the form of the number and shape of flow obstacles, the {\em problem physics} in the form of the Reynolds number, and the combination of the two.
Across all settings we find
that examples from lower-difficulty settings can significantly improve the performance on the associated hard test distribution.
Furthermore, this result holds for both supervised models like CNO and FFNO as well as the Poseidon family of state-of-the-art multi-physics-pretrained FMs.
This suggests that our observed performance gains are not only function of model class or model capacity 
but also a function of the quality and difficulty level of the training data distribution, as well as how it relates to the target distribution. 
In particular, we show that incorporating intermediate-difficulty examples has significant benefits.
Therefore, for a fixed computational budget, it may be more cost-effective to generate smaller number of high-quality moderately complex---i.e., intermediate---data, rather than relying on large volumes of simpler data.
Hence, our work suggests that future data-generation workflows for neural PDE solvers should take into account tradeoffs between the difficulty of generating low-to-medium-to-high complexity data and the potential benefits of harder-to-simulate data for learning that target distribution.\looseness-1

% We believe that 
% our insights provide interesting avenues for future work, including more formal methods for selecting datasets, more active data construction pipelines, and tests of the effects of data mixtures at scale.
%\vspace{-0.8em}
% \subsection{Broader impacts}
% This work highlights the importance of the data mixture for neural PDE solvers. 
% Since training a neural PDE solver requires classical simulation to pre-generate training data, our study can enable more energy-efficient strategies of doing so.
% This will in-turn enable more powerful PDE solvers that can provide a cheap alternative to larger classes of expensive classical simulation software.
% Our results can also shed insights into related applications of fluid dynamics, climate modeling and weather forecasting where getting such simulations is expensive.

\section*{Reproducibility Statement}
We took several steps to make our results easy to reproduce. Problem setups, difficulty axes, dataset sizes/splits, and the evaluation metric are specified in the main text (Secs.~\ref{sec:setup}–\ref{sec:experiments}), including the definition of nMAE in Eq.~\eqref{eq:nmae} and the exact target-and-mix protocols summarized in Figures 4-9. 
The full simulation pipeline for generating FPO and LDC datasets—covering domain construction, boundary conditions, Reynolds-number sampling, time scheduling, discretization schemes, solver settings, post-processing to a $128\times128$ grid, and the saved data format—is documented in the \textbf{Supplementary Material, App.~\ref{app:dataset}} (FPO in Sec.~\ref{app:fpo}, LDC in Sec.~\ref{app:ldc}, scheduling in Sec.~\ref{app:scheduling}, numerics in Sec.~\ref{app:numerics}, and data format in Sec.~\ref{app:data-format}; Table~\ref{tab:re-time-schedule}; Figs.~\ref{fig:alpha-sweep}, \ref{fig:fpo-snapshots}, \ref{fig:ldc-snapshots}). 
Model architectures, training/fine-tuning procedures, hyperparameters, and the compute environment are detailed in \textbf{Supplementary, App.~\ref{app:models}} (CNO in Sec.~\ref{app:cno}, F-FNO in Sec.~\ref{app:ffno}, Poseidon variants in Secs.~\ref{app:poseidon-t}–\ref{app:poseidon-l}), with training-time summaries in \textbf{App.~\ref{app:compute}} (Table~\ref{tab:training-time}). 
External corpora and out-of-distribution geometry experiments using FlowBench, and how they are combined with our pre-generated data, is described in Section.~\ref{sec:foundation}. 
We have released all pre-generated datasets and code to reproduce our results \href{https://github.com/Naman-Choudhary-AI-ML/pregenerating-pde}{at this url}.

\section*{Acknowledgments}

Mikhail Khodak acknowledges the resources of the Princeton Language and Intelligence unit of the Princeton AI Lab.

\bibliographystyle{plainnat}
\bibliography{refs}

@inproceedings{dauner2024resffno,
  title     = {Residual Factorized Fourier Neural Operator for Simulation of Three-Dimensional Turbulent Flows},
  author    = {Dauner, Maximilian and others},
  booktitle = {OpenReview preprint (ICLR submission) id: yGdoTL9g18},
  year      = {2024},
  url       = {https://openreview.net/forum?id=yGdoTL9g18}
}

@inproceedings{koehler2024apebench,
  title     = {A Benchmark for Autoregressive Neural Emulators of PDEs},
  author    = {Koehler, Frederic and others},
  booktitle = {NeurIPS 2024 Datasets and Benchmarks Track},
  year      = {2024},
  url       = {https://arxiv.org/abs/2411.00180}
}

@inproceedings{krishnapriyan2021characterizing,
  title        = {Characterizing possible failure modes in physics-informed neural networks},
  author       = {Krishnapriyan, Aditi S. and Gholami, Amir and Zhe, Shandian and Kirby, Robert M. and Mahoney, Michael W.},
  booktitle    = {Advances in Neural Information Processing Systems (NeurIPS)},
  volume       = {34},
  year         = {2021},
  url          = {https://proceedings.neurips.cc/paper/2021/hash/df438e5206f31600e6ae4af72f2725f1-Abstract.html},
  eprint       = {2109.01050},
  archivePrefix= {arXiv}
}

@article{raissi2019pinn,
title={Physics-informed neural networks: A deep learning framework for solving forward and inverse problems involving nonlinear partial differential equations},
author={Raissi, Maziar and Perdikaris, Paris and Karniadakis, George E.},
journal={Journal of Computational Physics},
volume={378},
pages={686--707},
year={2019}
}

@InProceedings{rotman2023semi,
  title = 	 {Semi-supervised learning of partial differential operators and dynamical flows},
  author =       {Rotman, Michael and Dekel, Amit and Ilan Ber, Ran and Wolf, Lior and Oz, Yaron},
  booktitle = 	 {Proceedings of the Thirty-Ninth Conference on Uncertainty in Artificial Intelligence},
  year = 	 {2023}
}

@article{karniadakis2021,
title={Physics-informed machine learning},
author={Karniadakis, George Em and Kevrekidis, Ioannis G. and Lu, Lu and Perdikaris, Paris and Wang, Sifan and Yang, Liu},
journal={Nature Reviews Physics},
volume={3},
number={6},
pages={422--440},
year={2021}
}

@article{lu2021deeponet,
title={Learning nonlinear operators via DeepONet based on the universal approximation theorem of operators},
author={Lu, Lu and Jin, Pengzhan and Pang, Guofei and Zhang, Zhongqiang and Karniadakis, George Em},
journal={Nature Machine Intelligence},
volume={3},
number={3},
pages={218--229},
year={2021}
}

@inproceedings{li2021fno,
title={Fourier Neural Operator for Parametric Partial Differential Equations},
author={Li, Zongyi and Kovachki, Nikola B. and Azizzadenesheli, Kamyar and Liu, Burigede and Bhattacharya, Kaushik and Stuart, Andrew and Anandkumar, Anima},
booktitle={International Conference on Learning Representations (ICLR)},
year={2021}
}

@article{kovachki2023,
title={Neural Operator: Learning Maps Between Function Spaces},
author={Kovachki, Nikola B. and Li, Zongyi and Liu, Burigede and Azizzadenesheli, Kamyar and Bhattacharya, Kaushik and Stuart, Andrew M. and Anandkumar, Anima},
journal={Journal of Machine Learning Research},
volume={24},
pages={1--79},
year={2023}
}

@inproceedings{brandstetter2022,
title={Message Passing Neural PDE Solvers},
author={Brandstetter, Johannes and Welling, Max and Worrall, Daniel E.},
booktitle={International Conference on Learning Representations (ICLR)},
year={2022}
}

@inproceedings{li2023geofno,
title={Fourier Neural Operator with Learned Deformations for PDEs on General Geometries},
author={Li, Zongyi and Gupta, Kamyar and Kovachki, Nikola B. and Azizzadenesheli, Kamyar and Anandkumar, Anima},
booktitle={Advances in Neural Information Processing Systems (NeurIPS)},
year={2023}
}

@article{brunton2020,
title={Machine Learning for Fluid Mechanics},
author={Brunton, Steven L. and Kutz, J. Nathan},
journal={Annual Review of Fluid Mechanics},
volume={52},
pages={477--508},
year={2020}
}

@article{tran2021factorized,
  title        = {Factorized Fourier Neural Operators},
  author       = {Tran, Anh and Mathews, Alexander and Xie, Lexing and Ong, Cheng Soon},
  journal      = {arXiv preprint arXiv:2111.13802},
  year         = {2021},
  eprint       = {2111.13802},
  archivePrefix= {arXiv},
  primaryClass = {cs.LG},
  url          = {https://arxiv.org/abs/2111.13802}
}

@misc{li2021fourierneuraloperatorparametric,
  title={Fourier Neural Operator for Parametric Partial Differential Equations}, 
  author={Zongyi Li and Nikola Kovachki and Kamyar Azizzadenesheli and Burigede Liu and Kaushik Bhattacharya and Andrew Stuart and Anima Anandkumar},
  year={2021},
  eprint={2010.08895},
  archivePrefix={arXiv},
  primaryClass={cs.LG},
  url={https://arxiv.org/abs/2010.08895}
}

@article{Lu_2021,
  title={Learning nonlinear operators via DeepONet based on the universal approximation theorem of operators},
  volume={3},
  ISSN={2522-5839},
  url={http://dx.doi.org/10.1038/s42256-021-00302-5},
  DOI={10.1038/s42256-021-00302-5},
  number={3},
  journal={Nature Machine Intelligence},
  publisher={Springer Science and Business Media LLC},
  author={Lu, Lu and Jin, Pengzhan and Pang, Guofei and Zhang, Zhongqiang and Karniadakis, George Em},
  year={2021},
  month=mar,
  pages={218--229}
}

@misc{guibas2022adaptivefourierneuraloperators,
  title={Adaptive Fourier Neural Operators: Efficient Token Mixers for Transformers}, 
  author={John Guibas and Morteza Mardani and Zongyi Li and Andrew Tao and Anima Anandkumar and Bryan Catanzaro},
  year={2022},
  eprint={2111.13587},
  archivePrefix={arXiv},
  primaryClass={cs.CV},
  url={https://arxiv.org/abs/2111.13587}
}

@misc{brandstetter2023messagepassingneuralpde,
  title={Message Passing Neural PDE Solvers}, 
  author={Johannes Brandstetter and Daniel Worrall and Max Welling},
  year={2023},
  eprint={2202.03376},
  archivePrefix={arXiv},
  primaryClass={cs.LG},
  url={https://arxiv.org/abs/2202.03376}
}

@misc{pathak2022fourcastnetglobaldatadrivenhighresolution,
  title={FourCastNet: A Global Data-driven High-resolution Weather Model using Adaptive Fourier Neural Operators}, 
  author={Jaideep Pathak and Shashank Subramanian and Peter Harrington and Sanjeev Raja and Ashesh Chattopadhyay and Morteza Mardani and Thorsten Kurth and David Hall and Zongyi Li and Kamyar Azizzadenesheli and Pedram Hassanzadeh and Karthik Kashinath and Animashree Anandkumar},
  year={2022},
  eprint={2202.11214},
  archivePrefix={arXiv},
  primaryClass={physics.ao-ph},
  url={https://arxiv.org/abs/2202.11214}
}

@misc{tali2024flowbenchlargescalebenchmark,
  title={FlowBench: A Large Scale Benchmark for Flow Simulation over Complex Geometries}, 
  author={Ronak Tali and Ali Rabeh and Cheng-Hau Yang and Mehdi Shadkhah and Samundra Karki and Abhisek Upadhyaya and Suriya Dhakshinamoorthy and Marjan Saadati and Soumik Sarkar and Adarsh Krishnamurthy and Chinmay Hegde and Aditya Balu and Baskar Ganapathysubramanian},
  year={2024},
  eprint={2409.18032},
  archivePrefix={arXiv},
  primaryClass={physics.flu-dyn},
  url={https://arxiv.org/abs/2409.18032}
}

@article{doi:10.1126/sciadv.aau6792,
  author    = {Jordan Hoffmann and Yohai Bar-Sinai and Lisa M. Lee and Jovana Andrejevic and Shruti Mishra and Shmuel M. Rubinstein and Chris H. Rycroft},
  title     = {Machine learning in a data-limited regime: Augmenting experiments with synthetic data uncovers order in crumpled sheets},
  journal   = {Science Advances},
  volume    = {5},
  number    = {4},
  pages     = {eaau6792},
  year      = {2019},
  doi       = {10.1126/sciadv.aau6792},
  url       = {https://www.science.org/doi/abs/10.1126/sciadv.aau6792}
}

@article{takamoto2022pdebench,
  title={PDEBench: An Extensive Benchmark for Scientific Machine Learning},
  author={Takamoto, Makoto and Praditia, Timothy and Leiteritz, Raphael and Others},
  journal={NeurIPS Datasets and Benchmarks},
  year={2022}
}

@article{ruiz2024benefits,
  title={On the benefits of memory for modeling time-dependent pdes},
  author={Ruiz, Ricardo Buitrago and Marwah, Tanya and Gu, Albert and Risteski, Andrej},
  journal={arXiv preprint arXiv:2409.02313},
  year={2024}
}

@article{ohana2024well,
  title={The well: a large-scale collection of diverse physics simulations for machine learning},
  author={Ohana, Ruben and McCabe, Michael and Meyer, Lucas and Morel, Rudy and Agocs, Fruzsina and Beneitez, Miguel and Berger, Marsha and Burkhart, Blakesly and Dalziel, Stuart and Fielding, Drummond and others},
  journal={Advances in Neural Information Processing Systems},
  volume={37},
  pages={44989--45037},
  year={2024}
}

@article{herde2024poseidon,
  title={Poseidon: Efficient foundation models for pdes},
  author={Herde, Maximilian and Raonic, Bogdan and Rohner, Tobias and K{\"a}ppeli, Roger and Molinaro, Roberto and de B{\'e}zenac, Emmanuel and Mishra, Siddhartha},
  journal={Advances in Neural Information Processing Systems},
  volume={37},
  pages={72525--72624},
  year={2024}
}

@article{hao2024dpot,
  title={Dpot: Auto-regressive denoising operator transformer for large-scale pde pre-training},
  author={Hao, Zhongkai and Su, Chang and Liu, Songming and Berner, Julius and Ying, Chengyang and Su, Hang and Anandkumar, Anima and Song, Jian and Zhu, Jun},
  journal={arXiv preprint arXiv:2403.03542},
  year={2024}
}

@article{shen2024ups,
  title={Ups: Towards foundation models for pde solving via cross-modal adaptation},
  author={Shen, Junhong and Marwah, Tanya and Talwalkar, Ameet},
  journal={arXiv e-prints},
  pages={arXiv--2403},
  year={2024}
}

@article{bruna2024neural,
  title={Neural Galerkin schemes with active learning for high-dimensional evolution equations},
  author={Bruna, Joan and Peherstorfer, Benjamin and Vanden-Eijnden, Eric},
  journal={Journal of Computational Physics},
  volume={496},
  pages={112588},
  year={2024},
  publisher={Elsevier}
}

@article{musekamp2024active,
  title={Active learning for neural pde solvers},
  author={Musekamp, Daniel and Kalimuthu, Marimuthu and Holzm{\"u}ller, David and Takamoto, Makoto and Niepert, Mathias},
  journal={arXiv preprint arXiv:2408.01536},
  year={2024}
}

@article{raonic2023convolutional,
  title={Convolutional neural operators for robust and accurate learning of PDEs},
  author={Raonic, Bogdan and Molinaro, Roberto and De Ryck, Tim and Rohner, Tobias and Bartolucci, Francesca and Alaifari, Rima and Mishra, Siddhartha and de B{\'e}zenac, Emmanuel},
  journal={Advances in Neural Information Processing Systems},
  volume={36},
  pages={77187--77200},
  year={2023}
}

@article{mccabe2023multiple,
  title={Multiple physics pretraining for physical surrogate models},
  author={McCabe, Michael and Blancard, Bruno R{\'e}galdo-Saint and Parker, Liam Holden and Ohana, Ruben and Cranmer, Miles and Bietti, Alberto and Eickenberg, Michael and Golkar, Siavash and Krawezik, Geraud and Lanusse, Francois and others},
  journal={arXiv preprint arXiv:2310.02994},
  year={2023}
}

@article{nadig2025consequences,
  title={Consequences of training data composition for deep learning models in single-cell biology},
  author={Nadig, Ajay and Thoutam, Akshaya and Hughes, Madeline and Gupta, Anay and Navia, Andrew W and Fusi, Nicolo and Raghavan, Srivatsan and Winter, Peter S and Amini, Ava P and Crawford, Lorin},
  journal={bioRxiv},
  pages={2025--02},
  year={2025},
  publisher={Cold Spring Harbor Laboratory}
}

@article{anil2022exploring,
  title={Exploring length generalization in large language models},
  author={Anil, Cem and Wu, Yuhuai and Andreassen, Anders and Lewkowycz, Aitor and Misra, Vedant and Ramasesh, Vinay and Slone, Ambrose and Gur-Ari, Guy and Dyer, Ethan and Neyshabur, Behnam},
  journal={Advances in Neural Information Processing Systems},
  volume={35},
  pages={38546--38556},
  year={2022}
}

@article{cho2024position,
  title={Position coupling: Improving length generalization of arithmetic transformers using task structure},
  author={Cho, Hanseul and Cha, Jaeyoung and Awasthi, Pranjal and Bhojanapalli, Srinadh and Gupta, Anupam and Yun, Chulhee},
  journal={arXiv preprint arXiv:2405.20671},
  year={2024}
}

@article{sun2024easy,
  title={Easy-to-hard generalization: Scalable alignment beyond human supervision},
  author={Sun, Zhiqing and Yu, Longhui and Shen, Yikang and Liu, Weiyang and Yang, Yiming and Welleck, Sean and Gan, Chuang},
  journal={arXiv preprint arXiv:2403.09472},
  year={2024}
}

@article{hase2024unreasonable,
  title={The unreasonable effectiveness of easy training data for hard tasks},
  author={Hase, Peter and Bansal, Mohit and Clark, Peter and Wiegreffe, Sarah},
  journal={arXiv preprint arXiv:2401.06751},
  year={2024}
}

@inproceedings{jasak2007openfoam,
  title={OpenFOAM: A C++ library for complex physics simulations},
  author={Jasak, Hrvoje and Jemcov, Aleksandar and Tukovic, Zeljko and others},
  booktitle={International workshop on coupled methods in numerical dynamics},
  volume={1000},
  pages={1--20},
  year={2007},
  organization={Dubrovnik, Croatia)}
}

@article{weller1998tensor,
  title={A tensorial approach to computational continuum mechanics using object-oriented techniques},
  author={Weller, Henry G and Tabor, Gavin and Jasak, Hrvoje and Fureby, Christer},
  journal={Computers in physics},
  volume={12},
  number={6},
  pages={620--631},
  year={1998},
  publisher={American Institute of Physics}
}

@phdthesis{jasak1996error,
  title={Error analysis and estimation for the finite volume method with applications to fluid flows},
  author={Jasak, Hrvoje},
  year={1996},
  school={Imperial College London, University of London}
}

%%%%%%%%%%%%%%%%%%%%%%%%%%%%%%%%%%%%%%%%%%%%%%%%%%%%%%%%%%%%

\appendix
\newpage
% !TEX root = main.tex

% Supplementary Material Table of Contents

% \textbf{Table of Contents}

% \begin{itemize}
%     \item A Experimental Setup
%     \begin{itemize}
%         \item A.1 Dataset Generation
%         \item A.2 Model Architectures
%         \begin{itemize}
%             \item A.2.1 Convolutional Neural Operator (CNO)
%             \item A.2.2 Poseidon Transformer (T)
%         \end{itemize}
%     \end{itemize}
%     \item B Additional Results
%     \item C Computational Resources
%     \item D Reproducibility Notes
% \end{itemize}

% \newpage

% !TEX root = main.tex
% \section*{Supplementary Material}
\vspace{-3mm}
\section{Related work}\label{sec:related}
\vspace{-2mm}
%
% \nb{Can we organize these contributions into categories using the paragraph environment? Overall, I think we need a better review of related work here. I would say we should have one paragraph on just general PINNs / neural operators, explaining their benefits and downfalls. Then maybe one on reducing the amount of data needed, combined datasets, active learning, foundationm mdoels, etc. Then one on how similar ideas have appeared for other modalities.}
%

% A primary goal of neural PDE solver research is to reduce scientists' dependence on classical numerical solvers and thus enable solutions to otherwise intractable problems.
% %
% %
% Many approaches take a model-driven perspective,
% focusing on developing architectures that can adapt to problem geometry or other aspects of specific 
% settings to reduce~\citep{li2023geometryinformedneuraloperatorlargescale, kochkov2021machine, marwah2023deep}
% There has also been significant work on learning methodologies that can make use of either 
% {\em active} in-domain sampling~\citep{bruna2024neural}
% or transfer from existing datasets~\citep{herde2024poseidon,shen2024ups,hao2024dpot, mccabe2023multiple}.

Two leading paradigms for learning PDE solutions are physics-informed neural networks~(PINNs) and neural operators. PINNs embed PDE residuals and boundary conditions into the loss, enabling mesh-free training, strong use of physics priors, and efficacy in small-data forward/inverse settings \citep{raissi2019pinn,karniadakis2021}. At the same time, training can involve challenging multi-term loss balancing and optimization stiffness~\citep{krishnapriyan2021characterizing}, 
with slower convergence on multi-scale or chaotic regimes (e.g., high Re turbulence) and sensitivity to complex geometries or boundary conditions \citep{karniadakis2021}. Neural operators (e.g., DeepONet, FNO) learn mappings between function spaces, providing amortized inference, cross-discretization/geometry generalization, and scalability via pretraining on synthetic data \citep{lu2021deeponet,li2021fno,kovachki2023}. Their performance, however, typically depends on substantial supervised datasets; robustness may be reduced under distribution shift across physics/geometry, and accuracy near shocks/discontinuities or conservation/stability guarantees may require additional structure and memory \citep{kovachki2023,brandstetter2022,ruiz2024benefits}. Geometry-aware operator variants (e.g., GeoFNO) enhance robustness on irregular domains yet still rely on curated simulation corpora \citep{li2023geofno}. Recent surveys synthesize these properties across PDE tasks, including turbulent flows~\citep{karniadakis2021,brunton2020}.\looseness-1

Unlike these efforts, we focus on the data itself, specifically on how generating better quality data may improve performance.
There exists some work in the sciences on augmenting scarce experimental datasets with abundant simulated data from simplified systems, e.g. \citet{doi:10.1126/sciadv.aau6792} demonstrated that combining simulated flat-folding patterns with limited experimental data enabled machine learning models to recover structure in crumpled sheets.
Similar studies are being conducted in biology, where recent work investigates the effects of training data composition on the performance of 
foundation models for single-cell
genomics~\citep{nadig2025consequences}.
We view our contribution as a more systematic study of how to generate and make use of data of different qualities.

Outside of PDEs, data difficulty has also been explored in other areas such as language modeling. 
For example, a major difficulty axis in natural language processing is context length, with several explorations of how to train models capable of solving long-context tasks without resorting to purely 
long-context training~\citep{anil2022exploring, cho2024position}.
Separately, our work is also related to work on {\em easy-to-hard generalization} 
in arithmetic reasoning tasks~\citep{sun2024easy,hase2024unreasonable}
Here, the past work has found that appropriately training the models on simpler tasks---for example simpler
math problems---leads to better performance on harder tasks.
Here a task's hardness is determined according to some human hardness measures, e.g. grade-level for STEM problems.
% A.2 Model Architecture
\section{Dataset Generation and Simulation Setup}
\label{app:dataset}

We generate two major datasets---Flow Past Object (FPO) and Lid-Driven Cavity (LDC)---to investigate the impact of domain complexity on the performance and generalization of neural PDE solvers. Each dataset contains three levels of difficulty: \textit{easy} (no obstacles), \textit{medium} (a single obstacle), and \textit{hard} (2–10 randomly placed obstacles). All simulations are run using \texttt{OpenFOAM}, a finite-volume CFD solver.

\subsection{FPO Domain}\label{app:fpo}
In the FPO setting, we simulate flow around one or more square obstacles within a $2 \times 2$ m rectangular domain using the \texttt{icoFoam} solver. The left boundary is treated as a velocity inlet, where we impose a parabolic inflow profile representative of fully developed laminar channel flow. The right boundary is set as a pressure outlet with fixed value, and the top and bottom boundaries are treated as no-slip walls. 

To define the parabolic inlet condition, we prescribe the horizontal velocity component $u(y)$ across the height $H = 2$ m of the domain using the analytical profile for plane Poiseuille flow:
\[
u(y) = 4 U_{\text{max}} \cdot \frac{y(H - y)}{H^2}, \quad y \in [0, H],
\]
where $U_{\text{max}}$ is the peak velocity occurring at the vertical midline ($y = H/2$). This ensures zero velocity at the top and bottom walls ($y = 0, H$) and a smooth parabolic profile across the inlet face.

Reynolds numbers are sampled from a truncated normal distribution $\mathcal{N}(5000, 2000^2)$ with support in $[100, 10000]$, and the corresponding $U_{\text{max}}$ is scaled to satisfy:
\[
\text{Re} = \frac{U_{\text{avg}} \cdot L}{\nu}, \quad \text{with } U_{\text{avg}} = \frac{2}{3} U_{\text{max}}, \quad L = 2 \text{ m}, \quad \nu = 1.5 \times 10^{-5} \text{ m}^2/\text{s},
\]
where $U_{\text{avg}}$ is the mean velocity of the parabolic profile. Solving for $U_{\text{max}}$ ensures consistency between the desired Reynolds number and the imposed inlet condition.

Obstacle configurations are generated by randomly placing between 2 and 10 square holes in the domain, using a rejection sampling algorithm to prevent overlap or boundary collision. For each simulation:
\begin{itemize}
    \item The geometry is procedurally constructed by modifying \texttt{blockMeshDict}, and mesh generation is handled via OpenFOAM's native utilities.
    \item The simulation duration is dynamically adjusted based on the sampled Reynolds number using a characteristic time scale, and outputs are recorded at 20 evenly spaced intervals.
    \item Velocity and pressure fields are post-processed using OpenFOAM utilities and interpolated onto a $128 \times 128$ uniform grid via barycentric interpolation.
\end{itemize}

\begin{figure}[!t]
    \centering
    \includegraphics[width=0.49\linewidth]{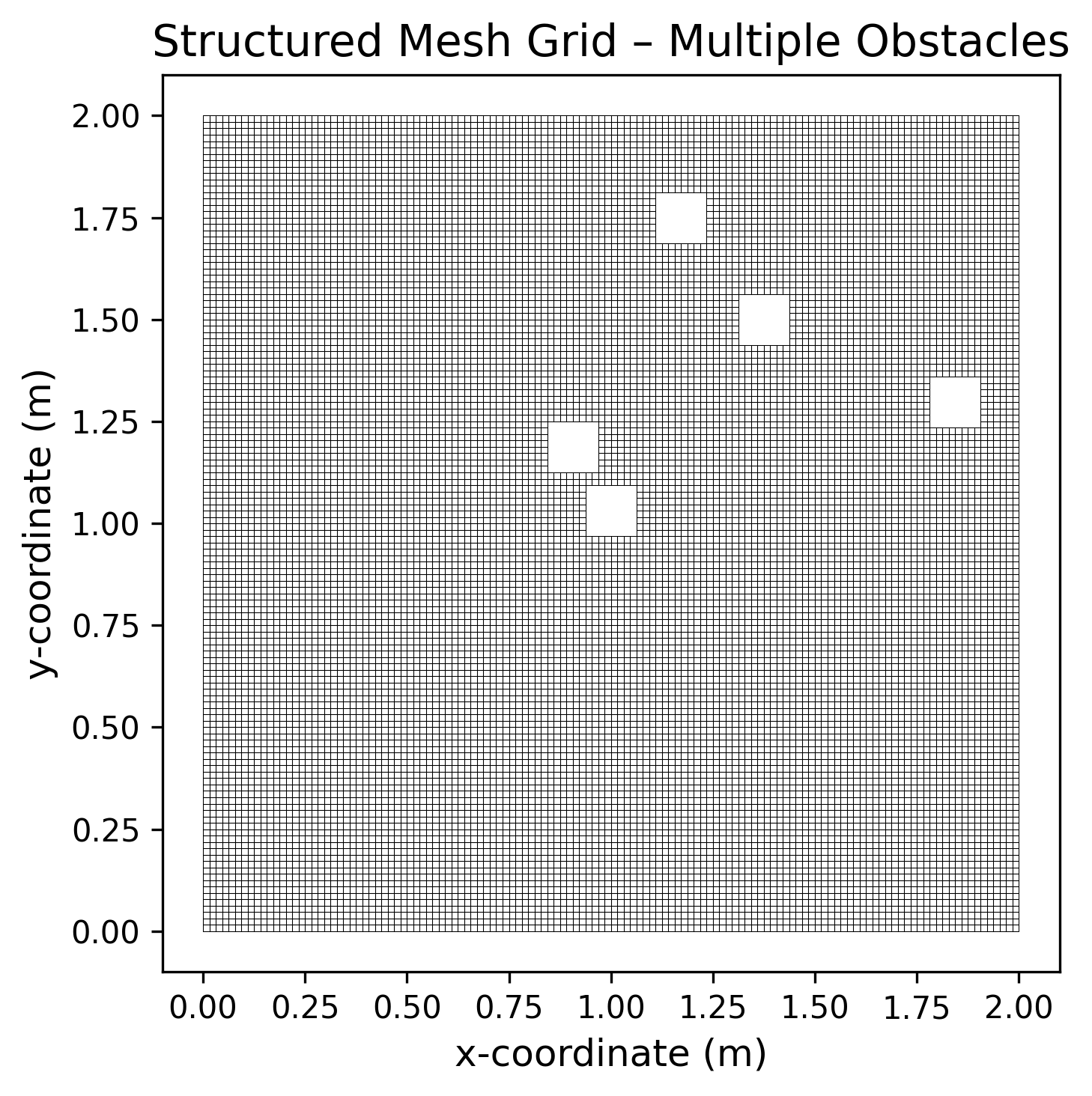}
    \includegraphics[width=0.49\linewidth]{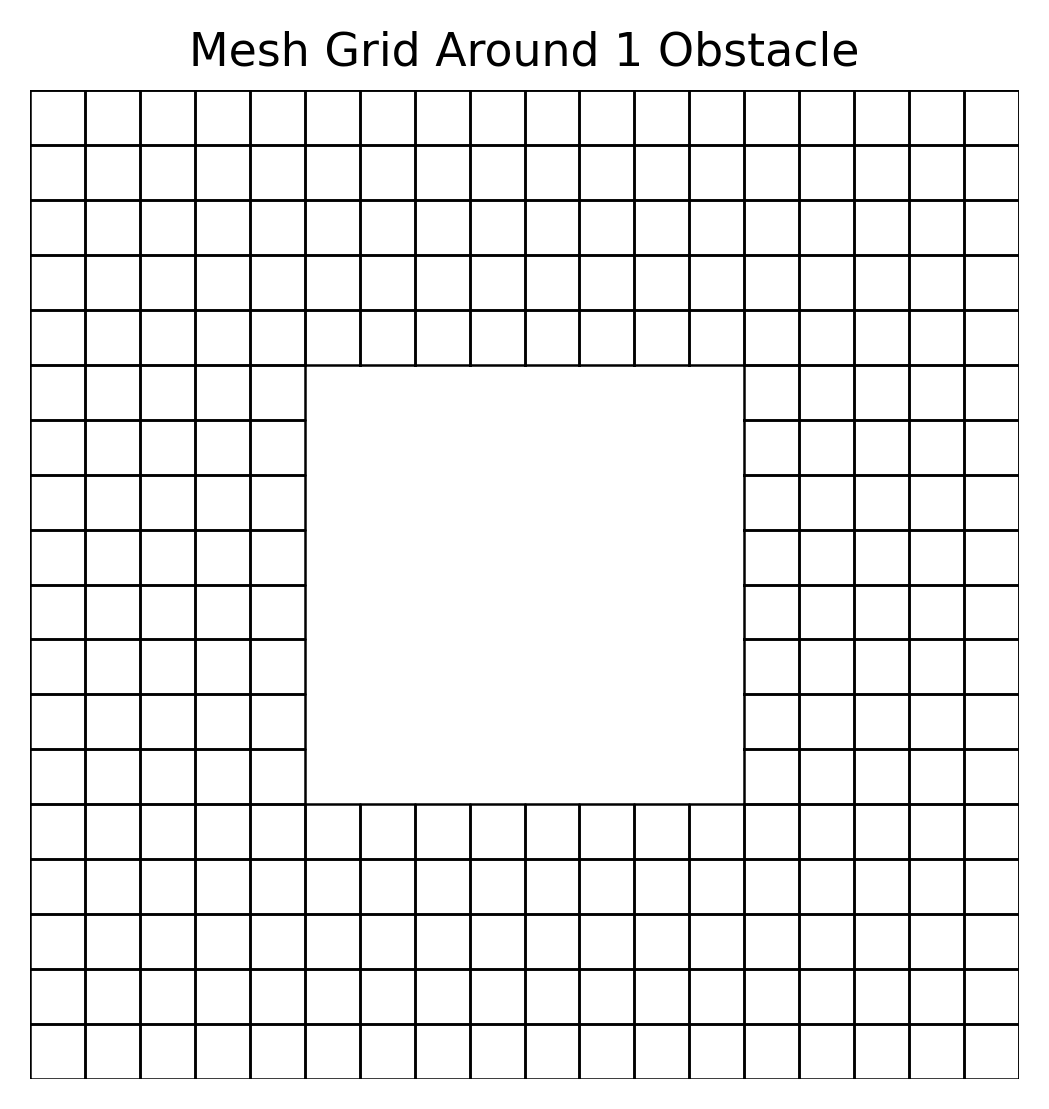}
    \caption{
    Structured mesh used in our multi obstacle setup. \textbf{Left}: full $2 \times 2$~m mesh domain. \textbf{Right}: zoom-in around one obstacle.
    }
    \label{fig:alpha-sweep}
\end{figure}

% \begin{figure}[!t]
%     \centering
    
%     % Easy (Velocity)
%     \begin{subfigure}[b]{0.9\textwidth}
%         \centering
%         \includegraphics[width=\textwidth]{Images/FPO_easy_vel.png}
%         \caption{\footnotesize Easy (Velocity)}
%         \label{fig:fpo-easy-vel}
%     \end{subfigure}
%     \vspace{1em}  % extra space between subfigures

%     % Intermediate (Velocity)
%     \begin{subfigure}[b]{0.9\textwidth}
%         \centering
%         \includegraphics[width=\textwidth]{Images/FPO_inter_vel.png}
%         \caption{\footnotesize Intermediate (Velocity)}
%         \label{fig:fpo-inter-vel}
%     \end{subfigure}
%     \vspace{1em}

%     % Complex (Velocity)
%     \begin{subfigure}[b]{0.9\textwidth}
%         \centering
%         \includegraphics[width=\textwidth]{Images/FPO_complex_vel.png}
%         \caption{\footnotesize Complex (Velocity)}
%         \label{fig:fpo-complex-vel}
%     \end{subfigure}

%     \caption{\textbf{FPO Flow Fields (Velocity).} 
%     Top row shows velocity fields for easy, intermediate, and complex setups;  
%     White areas represent the square obstacles (holes) in the domain.}
%     \label{fig:fpo-snapshots}
% \end{figure}

\begin{figure}[!t]
    \centering
        \centering
        \includegraphics[width=\textwidth]{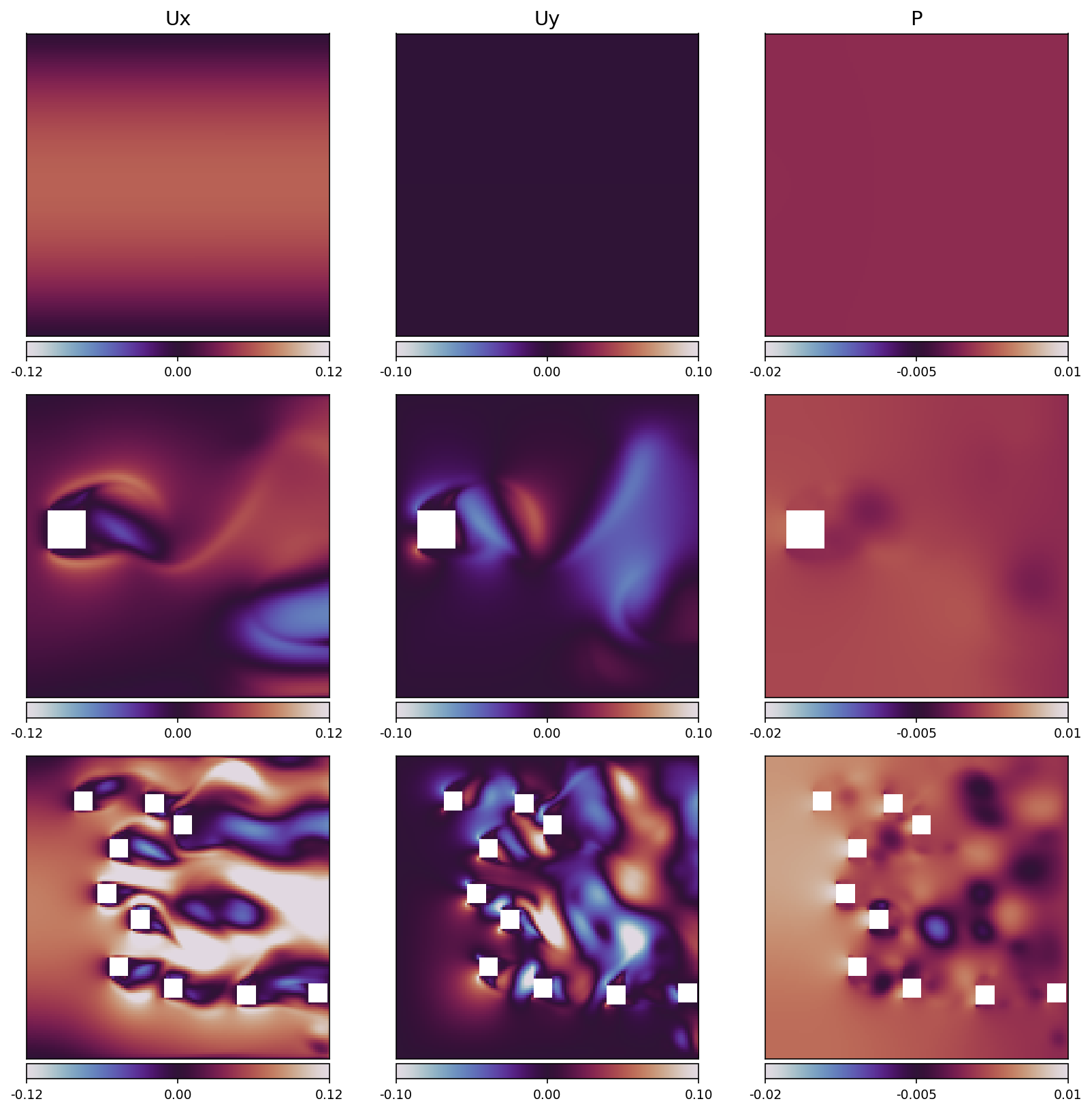}
        \label{fig:fpo-easy-vel}
    \caption{\textbf{FPO Flow Fields (Velocity).} 
    Top row shows velocity and relative pressure fields for easy, medium, and complex setups;  
    White areas represent the square obstacles (holes) in the domain. \textbf{Note:} The visualized pressure represents relative (gauge) pressure and may be negative, as the incompressible Navier-Stokes equations depend only on pressure gradients rather than absolute pressure.}
    \label{fig:fpo-snapshots}
\end{figure}

\subsection{LDC Domain}\label{app:ldc}
In the LDC setting, fluid flows in a closed $2 \times 2$ m cavity with a moving top wall. We again use the \texttt{icoFoam} solver with zero velocity on side and bottom walls and a parabolic profile imposed on the top wall. The top-wall velocity is scaled to match a target Reynolds number:
\[
U_{\text{max}} = \frac{\text{Re} \cdot \nu}{L}, \quad \text{with } \nu = 1.5 \times 10^{-5} \text{ m}^2/\text{s}, L = 2 \text{ m}.
\]

\subsection{Flow Development Scheduling.}\label{app:scheduling}
To ensure that each simulation reaches a fully developed state before data is recorded, we adaptively determine the simulation end time based on the sampled Reynolds number $\text{Re}$. This is critical in both the FPO and LDC domains, where flow transients can vary significantly with $\text{Re}$, and premature truncation would lead to incomplete or biased solution fields.

We define a piecewise scheduling rule that maps the Reynolds number to a simulation end time $T_{\text{end}}$ via either a linear scaling or a constant duration, depending on the flow regime. For moderate to high $\text{Re}$ values, we employ the characteristic viscous diffusion time scale:
\[
t_{\text{nd}} = \frac{L^2}{\nu \cdot \text{Re}},
\]
where $L = 2$ m is the characteristic length of the domain, and $\nu = 1.5 \times 10^{-5}$ m\textsuperscript{2}/s is the kinematic viscosity of the fluid. The total simulation time is then computed as:
\[
T_{\text{end}} = \gamma \cdot t_{\text{nd}},
\]
where $\gamma$ is a multiplicative factor that increases with $\text{Re}$ to accommodate longer transients associated with inertial effects. For very low Reynolds numbers ($\text{Re} < 100$), where steady-state is approached slowly and viscous effects dominate, we assign a fixed total duration of 2700 seconds.

The multiplicative factors $\gamma$ are manually calibrated for different $\text{Re}$ ranges as shown below:

\begin{table}[ht]
\centering
\begin{tabular}{ccc}
\toprule
Re Range & $\gamma$ & $T_{\text{end}}$ Formula \\
\midrule
$5000$–$10000$   & 40 & $T_{\text{end}} = 40 \cdot t_{\text{nd}}$ \\
$4000$–$5000$    & 30 & $T_{\text{end}} = 30 \cdot t_{\text{nd}}$ \\
$2500$–$4000$    & 20 & $T_{\text{end}} = 20 \cdot t_{\text{nd}}$ \\
$1000$–$2500$    & 10 & $T_{\text{end}} = 10 \cdot t_{\text{nd}}$ \\
$500$–$1000$     & 5  & $T_{\text{end}} = 5 \cdot t_{\text{nd}}$ \\
$400$–$500$      & 4  & $T_{\text{end}} = 4 \cdot t_{\text{nd}}$ \\
$300$–$400$      & 3  & $T_{\text{end}} = 3 \cdot t_{\text{nd}}$ \\
$200$–$300$      & 2  & $T_{\text{end}} = 2 \cdot t_{\text{nd}}$ \\
$100$–$200$      & 1  & $T_{\text{end}} = 1 \cdot t_{\text{nd}}$ \\
$10$–$100$       & -- & $T_{\text{end}} = 2700\ \text{s (fixed)}$ \\
\bottomrule
\end{tabular}
\caption{Reynolds-number-dependent scheduling of simulation end time.}
\label{tab:re-time-schedule}
\end{table}

The computed $T_{\text{end}}$ is rounded up to the nearest hundred and used to configure the \texttt{controlDict} file for each simulation. The write interval is also dynamically selected to yield 20 evenly spaced output frames, ensuring consistent temporal sampling across all Reynolds number regimes. This scheduling mechanism guarantees physically meaningful and temporally aligned datasets, while avoiding wasted computation for low-Re flows or premature termination for higher-Re flows.

\begin{figure}[!t]
    \centering
        \includegraphics[width=\textwidth]{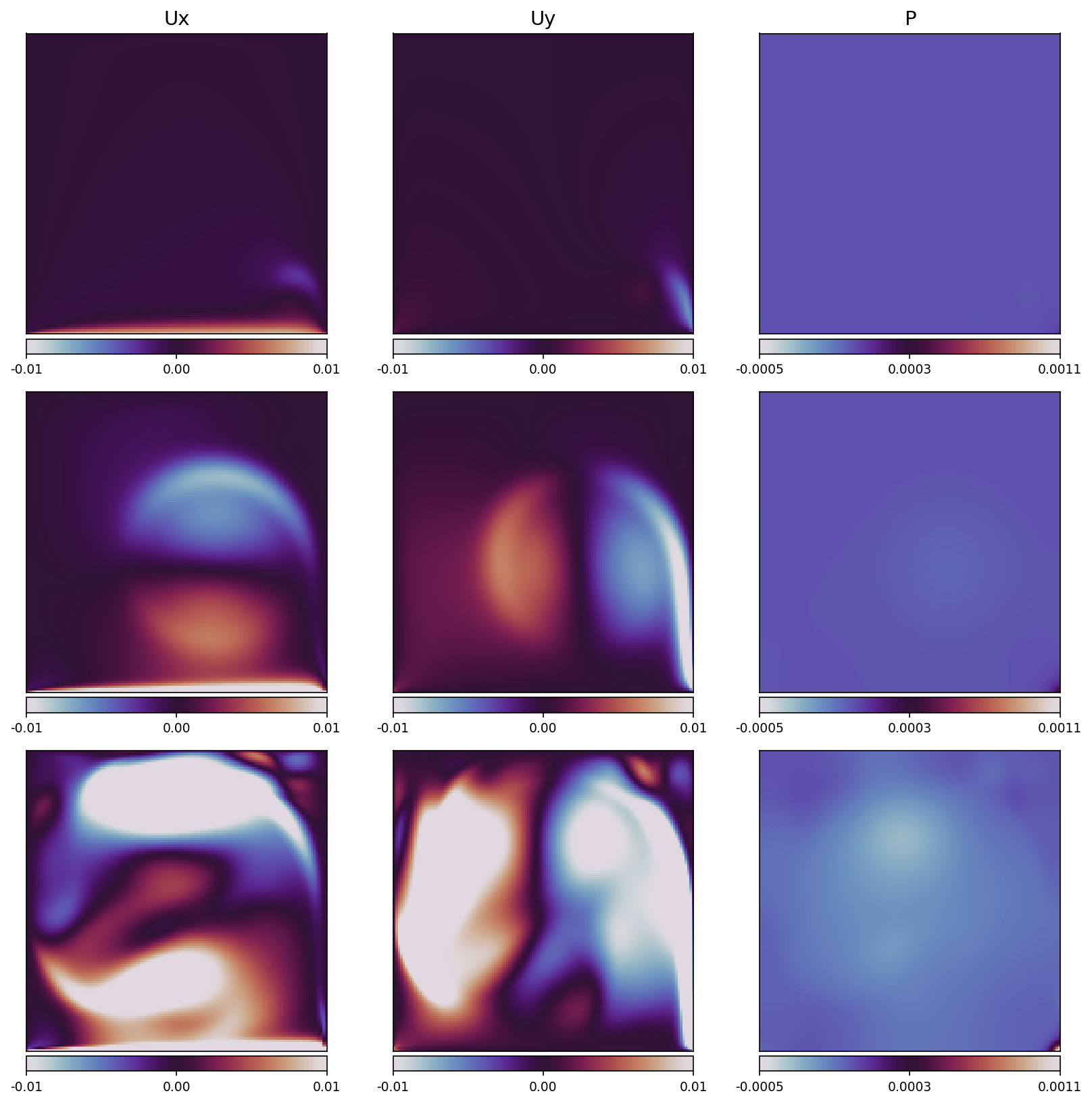}
    \caption{\textbf{LDC Flow Fields (Velocity).} 
    Top row: velocity and relative pressure fields for easy, medium, and complex setups. Note that pressure is shown as relative (gauge) pressure, which allows for negative values consistent with incompressible flow formulation.}
    \label{fig:ldc-snapshots}
\end{figure}
\subsection{Numerical Discretization and Solver Configuration}\label{app:numerics}

All simulations are performed using OpenFOAM v2406 to numerically solve the incompressible Navier–Stokes equations. To ensure stable and accurate data generation across diverse Reynolds numbers and geometric complexities, we adopt a consistent finite-volume setup for time integration, spatial discretization, and linear solver configurations.

\vspace{1mm}
\textbf{Governing Equations.}
We solve the incompressible Navier–Stokes system in the exact form used in the main paper:
\begin{equation}
    \partial_t\*u+(\*u\cdot\nabla)\*u+\nabla p=\nu\Delta\*u
    \qquad\textrm{and}\qquad
    \nabla\cdot\*u=0
\end{equation}
where $\*u(\*x,t)$ is the velocity field, $p$ is the (kinematic) pressure, and $\nu$ is the kinematic viscosity.

\subsubsection{Time Discretization}
To maintain numerical robustness at higher Reynolds numbers and small time steps, we discretize the temporal derivative using a first-order implicit backward-Euler scheme:
\[
\frac{\partial \mathbf{U}}{\partial t} \approx \frac{\mathbf{U}^{n+1} - \mathbf{U}^{n}}{\Delta t}.
\]
This choice offers unconditional stability and aligns with OpenFOAM’s standard transient solvers.

\subsubsection{Spatial Derivatives}
All spatial derivatives are evaluated using the finite-volume method with second-order accurate schemes:
\begin{itemize}
    \item \textbf{Gradient terms} such as $\nabla p$ and $\nabla \mathbf{U}$ use central differencing:
    \[
    \nabla \phi \approx \texttt{Gauss linear},
    \]
    which preserves smooth fields with low numerical diffusion.

    \item \textbf{Convective fluxes}, dominant at higher Reynolds numbers, use an upwind-biased linear scheme with gradient reconstruction:
    \[
    \nabla \cdot (\phi \mathbf{U}) \approx \texttt{Gauss linearUpwind grad(U)},
    \]
    balancing stability with second-order accuracy, especially near obstacles where steep gradients occur.

    \item \textbf{Diffusive terms} (Laplacians) use:
    \[
    \nabla^2 \phi \approx \texttt{Gauss linear orthogonal},
    \]
    appropriate for our structured Cartesian grids.\footnote{If mild non-orthogonality appears, \texttt{Gauss linear corrected} is a safe alternative.}
\end{itemize}

\subsubsection{Interpolation and Surface Gradients}
Cell-face values are interpolated linearly:
\[
\phi_f \approx \texttt{linear}(\phi),
\]
and surface-normal gradients use the \texttt{orthogonal} scheme, leveraging the grid’s structured nature.

\subsubsection{Linear Solvers}
The momentum and pressure equations are solved using efficient iterative solvers:
\begin{itemize}
  \item \textbf{Pressure} ($p$): PCG (Preconditioned Conjugate Gradient) with DIC (Diagonal-based Incomplete Cholesky) preconditioning.
  \item \textbf{Velocity} ($\mathbf{U}$): \texttt{smoothSolver} with symmetric Gauss–Seidel smoothing.
\end{itemize}
Per-equation tolerances are:
\begin{align*}
\text{Pressure:}&\quad \texttt{tolerance} = 10^{-6}, \quad \texttt{relTol} = 0.05 \ \text{(final: 0)} \\
\text{Velocity:}&\quad \texttt{tolerance} = 10^{-5}, \quad \texttt{relTol} = 0 \ \text{(final)}
\end{align*}

\subsubsection{Design Motivation}
This configuration follows established best practices in the OpenFOAM ecosystem and prior simulation-driven ML benchmarks, ensuring numerical stability and physical realism across a wide range of Reynolds numbers. We adopt \texttt{Gauss linear} for gradients and diffusive terms to preserve smoothness on structured grids~\cite{weller1998tensor}, and \texttt{linearUpwind grad(U)} to balance accuracy and robustness in the presence of sharp gradients and internal obstacles~\cite{jasak1996error}. The backward-Euler time integration and implicit solvers align with standard OpenFOAM settings for incompressible flows and are widely used in both industrial and academic studies~\cite{jasak2007openfoam}.

\subsubsection{Simulation Pipeline (Transient)}
We automate data generation via modular Python scripts for both FPO and LDC:
\begin{enumerate}
    \item \textbf{Domain Construction:} Randomized obstacle positions are sampled; a mesh is constructed via a modified \texttt{blockMeshDict}.
    \item \textbf{Velocity and Controls:} Boundary velocity profiles and run duration are computed from the sampled Reynolds number.
    \item \textbf{Simulation Execution:} The case is solved using \texttt{icoFoam}; fields are written at fixed intervals to yield $20$ timesteps.
    \item \textbf{Postprocessing:} Velocity and pressure fields are parsed and projected onto a $128 \times 128$ regular grid.
    \item \textbf{Geometry Encoding:} Each grid cell includes a binary mask (fluid vs.\ obstacle) and a signed distance field (SDF) via an Euclidean distance transform.
\end{enumerate}

\subsubsection{Randomized Obstacle Generation}
We construct domains with multiple internal holes via \texttt{blockMeshDict}:
\begin{itemize}
\item \emph{Random Hole Placement:} Sample $n\!\in\![2,10]$ axis-aligned rectangles $\{x,y,w,h\}$ strictly within $[0,2]\times[0,2]$ (optionally enforcing non-overlap).
\item \emph{Block Decomposition:} Subdivide a structured Cartesian grid; cells lying entirely inside holes are removed. Faces adjoining missing cells become boundary patches \texttt{hole1}, \texttt{hole2}, \ldots
\item \emph{Boundary Patches:} The outer walls (including the moving lid for LDC) and the hole patches are set as no-slip walls (\(\mathbf{U}=\mathbf{0}\)) unless the experiment specifies inlets/outlets (FPO).
\end{itemize}

\subsubsection{Boundary Conditions and Reynolds Number}
For LDC, the top-wall velocity $U_{\text{lid}}$ is set to match a target Reynolds number:
\[
\mathrm{Re} \;=\; \frac{U_{\text{lid}}\, L}{\nu}
\quad\Longrightarrow\quad
U_{\text{lid}} \;=\; \frac{\mathrm{Re}\,\nu}{L}.
\]
For FPO, inlet speed is set analogously; outlets use zero-gradient pressure and velocity conditions consistent with standard setups.

\subsubsection{Data Format}\label{app:data-format}
\begin{figure}[t]
    \centering
    \begin{subfigure}[t]{0.48\linewidth}
        \centering
        \includegraphics[width=\linewidth]{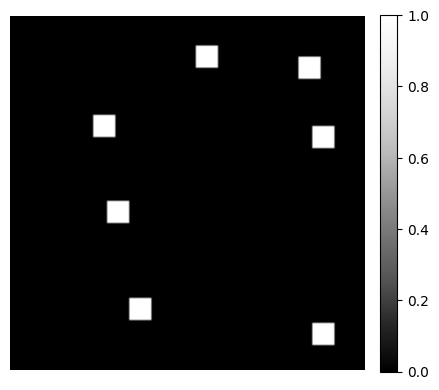}
        \caption{\footnotesize Binary mask: white = hole, black = fluid.}
        \label{fig:mask-research}
    \end{subfigure}
    \hfill
    \begin{subfigure}[t]{0.48\linewidth}
        \centering
        \includegraphics[width=\linewidth]{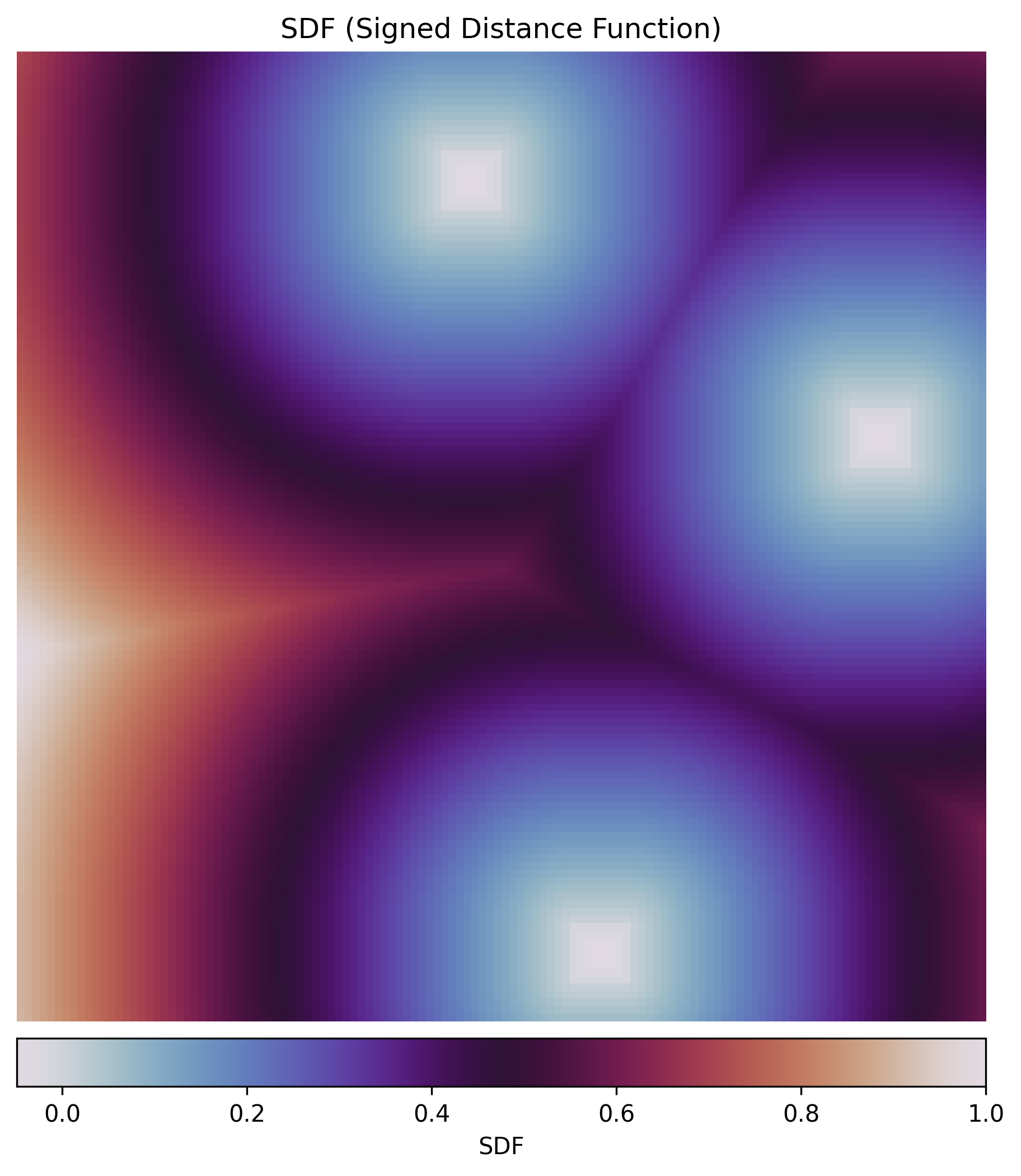}
        \caption{\footnotesize Signed distance field (SDF) corresponding to the mask.}
        \label{fig:sdf-research}
    \end{subfigure}
    \caption{\textbf{Geometry encodings.} Visualization of the binary mask and corresponding SDF used to encode obstacle geometry.}
    \label{fig:supp-fpo-example}
\end{figure}

Each trajectory is stored as a NumPy array with shape \texttt{(20, 128, 128, 6)}, containing six channels: horizontal velocity $u$, vertical velocity $v$, pressure $p$, normalized Reynolds number $\widehat{\mathrm{Re}}$, binary mask, and SDF. A representative visualization is shown in Figure~\ref{fig:supp-fpo-example}.

\section{Model Architectures}\label{app:models}

In this section, we provide implementation and training details for the neural operator models evaluated in our study. We focus on two classes of models: (1) the Convolutional Neural Operator (CNO)~\citep{raonic2023convolutional}, which is trained from scratch, and (2) Poseidon-T~\citep{herde2024poseidon}, a pretrained transformer-based model fine-tuned on our downstream task. These models differ significantly in their architectural design, parameterization, and training strategy.

All models are trained to predict velocity and pressure fields for 2D incompressible Navier-Stokes simulations. Given an input-output pair $(u_t, u_{t+1})$, where $u_t \in \mathbb{R}^{C \times H \times W}$ denotes the flow variables at timestep $t$ and $C$ is the number of channels, models are trained to minimize the relative $\ell_1$ loss:
\begin{equation}
\mathcal{L}(u_t, u_{t+1}) = \frac{\| u_{t+1} - \hat{u}_{t+1} \|_1}{\| u_{t+1} \|_1 + \epsilon},
\end{equation}
where $\hat{u}_{t+1}$ is the model prediction and $\epsilon = 10^{-10}$ ensures numerical stability.

\subsection{Convolutional Neural Operator (CNO)}\label{app:cno}

The Convolutional Neural Operator (CNO) used in our experiments is based on the architecture proposed by \citet{raonic2023convolutional}, with several adjustments tailored to our time-dependent Navier-Stokes flows. The model is implemented in PyTorch Lightning and trained in an autoregressive supervised setting. 

The model is structured as an encoder-decoder network with residual blocks, optional attention in the bottleneck, and optional time-conditioning via instance normalization. Key components of the architecture include:

\begin{itemize}
    \item \textbf{Encoder and Decoder:} The encoder has $N_\text{layers} = 4$ downsampling stages with a channel multiplier of 32. The decoder mirrors this with corresponding upsampling layers. Each stage consists of a combination of convolutional and spectral convolution blocks.
    \item \textbf{Residual Blocks:} The encoder and bottleneck include $N_\text{res} = 8$ residual blocks per level and $N_\text{res\_neck} = 8$ blocks in the bottleneck.
    \item \textbf{Normalization:} We use instance normalization (specified via $\texttt{nl\_dim} = [2, 3]$), conditional on the input timestep.
    \item \textbf{Time Conditioning:} The model is trained with $\texttt{is\_time} = 1$ to incorporate the current timestep as an additional input feature.
    \item \textbf{Grid Resolution:} All experiments use a spatial resolution of $128 \times 128$.
    \item \textbf{Loss Function:} We use a normalized relative $\ell_1$ loss, computed per-sample and averaged over the batch.
    \item \textbf{Autoregressive Training:} The CNO is trained in an autoregressive supervised manner using transitions of the form $t \rightarrow t{+}1$ only. We set $\texttt{allowed} = \texttt{'one'}$ in the training configuration, restricting the training to adjacent timestep pairs.
    \item \textbf{Other Details:} Training used a batch size of 32, learning rate of $7.5 \times 10^{-4}$, step learning-rate schedule over \textbf{400 epochs}, and weight decay of $1 \times 10^{-6}$.
\end{itemize}
0.00075
The model contains approximately 18 million parameters and does not utilize attention in our setup.

\subsection{Factorized Fourier Neural Operator (F-FNO)}\label{app:ffno}

The Factorized Fourier Neural Operator (F-FNO) used in our experiments is based on the architecture proposed by \citet{tran2021factorized}, designed for efficient learned simulation of PDEs. The model is implemented in PyTorch Lightning and trained in an autoregressive one-step prediction setting. It consists of a deep sequence of Fourier operator layers with factorized spectral convolutions and improved residual connections, which allow stable training even at greater network depths than the original FNO
. Key components of the architecture include:

\begin{itemize}
\item \textbf{Network Depth and Parameters:} We deploy a 5-layer F-FNO, following the high-capacity configuration from the original paper.

. This is an order of magnitude fewer parameters than a comparable geo-FNO model, despite the increased depth, due to the factorized kernel representation.
\item \textbf{Spectral Layers:} Each layer applies a separable Fourier convolution on the input features, factorizing the transform over each spatial dimension
. In practice, we drop half of the higher-frequency Fourier modes in each layer to reduce computational cost (e.g., on a $64\times 64$ grid we keep only the top 16 modes per dimension)
. The retained frequency components serve as learned global convolution kernels applied via inverse FFT.
\item \textbf{Feedforward Block:} After the spectral convolution, each layer includes a two-layer feed-forward network (pointwise MLP) operating in the physical space
. We use ReLU activations in these feed-forward layers
. This pointwise MLP mixes features per grid location and is analogous to the transformer’s MLP block, providing non-linear coupling of the channels.
\item \textbf{Residual Connections:} A skip connection adds each layer’s input to its output after the non-linear feed-forward stage
. This post-activation residual design preserves more of the original signal and stabilizes gradient flow in deep stacks
, enabling the F-FNO to converge even with dozens of layers (where the original FNO would diverge at $\geq 12$ layers
).
\item \textbf{Coordinate Encoding:} Following \citet{tran2021factorized}, we augment the input with explicit spatial coordinate channels. Each input field is concatenated with its normalized $x$ and $y$ coordinates (as two additional channels), providing a positional encoding that consistently improves accuracy
. The Fourier layers inherently utilize absolute positions (through the grid indices in the transform), while the feed-forward layers benefit from the coordinate features to capture location-dependent effects
.
\item \textbf{Autoregressive Training:} We train the F-FNO in a one-step-ahead prediction manner. The model uses only the current state as input to predict the next state, enforcing a first-order Markov assumption (no multi-step history)
. We employ teacher forcing during training, i.e. at each training step the ground-truth state at time $t$ is provided as input to predict $t{+}1$
. This approach avoids the need to unroll long sequences during training and was found to improve stability and accuracy.
\item \textbf{Input Normalization and Noise:} We apply per-variable normalization to input fields and add a small Gaussian noise perturbation during training
. These techniques, recommended by \citet{tran2021factorized}, act as regularization and help prevent training instabilities (we observed that without the added noise, the model’s validation loss could sometimes spike early in training
).
\item \textbf{Training Setup:} The F-FNO is trained with a batch size of 16, using a learning rate of $5\times 10^{-5}$ and a cosine annealing schedule (no restarts) over \textbf{400 epochs}, along with a weight decay of $1\times 10^{-6}$. These hyperparameters match those used for our other models to ensure a fair comparison. We did not employ the optional weight-sharing of Fourier weights across layers in our configuration, as it has minimal impact on performance at this depth
.
\end{itemize}

\subsection{Poseidon-T}\label{app:poseidon-t}

We evaluate Poseidon-T using the pretrained model checkpoints provided by \citet{herde2024poseidon}, available on Hugging Face.\footnote{\url{https://huggingface.co/camlab-ethz/Poseidon-T}} We perform fine-tuning on our custom datasets without any further pretraining.

The architecture follows a SwinUNet-based transformer backbone with hierarchical attention and patch embeddings. We retain the pretrained model structure and only update weights via supervised autoregressive finetuning. Key configuration details include:

\begin{itemize}
    \item \textbf{Backbone:} SwinUNet with hierarchical attention and window-based self-attention.
    \item \textbf{Variant:} We use Poseidon-\textbf{T}, which has a base embedding dimension of 48, depths $[4, 4, 4, 4]$, and patch size 4.
    \item \textbf{Resolution:} All inputs are processed at $128 \times 128$ resolution.
    \item \textbf{Training Setup:} Fine-tuning is performed for 100 epochs with batch size 16, weight decay of $1 \times 10^{-6}$, and cosine learning rate schedule starting from $5 \times 10^{-5}$.
    \item \textbf{Loss Function:} We use a normalized relative $\ell_1$ loss, computed per-sample and averaged over the batch.
\end{itemize}

Only the decoder and time-conditioning layers are updated during fine-tuning. The rest of the model remains unchanged from the pretrained checkpoint.

\subsection{Poseidon-B}\label{app:poseidon-b}

We evaluate Poseidon-B using the pretrained model checkpoint provided by \citet{herde2024poseidon}, available on Hugging Face.\footnote{\url{https://huggingface.co/camlab-ethz/Poseidon-B}} We perform fine-tuning on our custom datasets without any further pretraining. The architecture mirrors Poseidon-T’s setup, following a SwinUNet-based transformer backbone with hierarchical (U-Net style) multiscale attention and window-based self-attention. We retain the pretrained model structure and update weights via supervised autoregressive fine-tuning. Key configuration details include:
\begin{itemize}
    \item \textbf{Backbone:} SwinUNet with hierarchical attention (patch merging/expansion) and windowed self-attention.
    \item \textbf{Variant:} Poseidon-\textbf{B}, base embedding dimension $96$, depths $[8,\,8,\,8,\,8]$ (eight SwinV2 transformer blocks per level), patch size $4$.
    \item \textbf{Resolution:} All inputs are processed at $128 \times 128$ resolution (matching the pretraining grid size).\footnote{\url{https://huggingface.co/camlab-ethz}}
    \item \textbf{Training setup:} $100$ epochs, batch size $16$, weight decay $1\times10^{-6}$, cosine learning-rate schedule starting from $5\times10^{-5}$.
    \item \textbf{Loss:} Normalized relative $\ell_1$ loss, computed per-sample and averaged over the batch.
    \item \textbf{Fine-tuned parameters:} Only the decoder and time-conditioning layers are updated; all other weights remain frozen from the pretrained checkpoint.
\end{itemize}

\subsection{Poseidon-L}\label{app:poseidon-l}

We evaluate Poseidon-L using the pretrained model checkpoint provided by \citet{herde2024poseidon}, available on Hugging Face.\footnote{\url{https://huggingface.co/camlab-ethz/Poseidon-L}} We fine-tune this model on our custom datasets with no additional pretraining. The architecture is identical to the other Poseidon variants, employing the same SwinUNet-style transformer backbone with hierarchical multiscale attention and window-based (shifted-window) self-attention. We preserve the original model architecture and learn weights via supervised autoregressive fine-tuning. Key configuration details include:
\begin{itemize}
    \item \textbf{Backbone:} SwinUNet with hierarchical attention and window-based self-attention (shifted-window mechanism).
    \item \textbf{Variant:} Poseidon-\textbf{L}, base embedding dimension $192$, depths $[8,\,8,\,8,\,8]$ (eight SwinV2 transformer blocks at each level), patch size $4$.
    \item \textbf{Resolution:} All inputs are handled at $128 \times 128$ resolution, consistent with the model’s pretraining grid size.\footnote{\url{https://huggingface.co/camlab-ethz}}
    \item \textbf{Training setup:} $100$ epochs, batch size $16$, weight decay $1\times10^{-6}$, cosine learning-rate schedule starting from $5\times10^{-5}$.
    \item \textbf{Loss:} Normalized relative $\ell_1$ loss, computed per-sample and averaged over the batch.
    \item \textbf{Fine-tuned parameters:} Only the decoder and time-conditioning (time-encoded layer normalization) layers are updated during fine-tuning; all other weights are frozen to their pretrained values.
\end{itemize}

% All experiments were run using a single GPU with no mixed precision or distributed training. 
\subsection{Computational Resources}\label{app:compute}

We report compute details and training runtimes for the experiments conducted in this study. All models were trained on a single GPU using PyTorch with SLURM-based scheduling. The Convolutional Neural Operator (CNO) models were trained from scratch for 400 epochs, while Poseidon-T/B/L models were fine-tuned for 200 epochs from publicly released checkpoints.

CNO models were trained from scratch while Poseidon models were fine-tuned on a single NVIDIA L40S GPU on the Babel cluster. For autoregressive training with $t \rightarrow t{+}1$ supervision, training time scaled approximately with the number of trajectories. Table~\ref{tab:training-time} summarizes the approximate training time for varying dataset sizes. For completeness, we include FFNO (trained from scratch) and additional Poseidon variants (B and L) alongside Poseidon-T. FFNO and CNO were trained from scratch; all Poseidon variants (T/B/L) were fine-tuned.

\begin{table}[ht]
\centering
\caption{Approximate training durations with increasing number of training trajectories on a single NVIDIA L40S GPU. FFNO and CNO trained from scratch; Poseidon variants fine-tuned.}
\label{tab:training-time}
\begin{tabular}{lccccc}
\toprule
\textbf{Training Trajectories} & \textbf{CNO} & \textbf{FFNO} & \textbf{Poseidon-T} & \textbf{Poseidon-B} & \textbf{Poseidon-L} \\
\midrule
200  & 3h 30m & 1h      & 2h   & 6h 40m & -- \\
400  & 6h 48m & 7h      & 3h   & 11h 30m & -- \\
800  & 7h 55m & 1d 13h  & 9h   & 19h & 1d 3h \\
1600 & 1d 4h  & 1d 21h  & 18h  & 1d 16h & -- \\
\bottomrule
\end{tabular}
\end{table}

All experiments were run using a single GPU with no mixed precision or distributed training.

\subsection{Training Convergence Analysis}\label{app:convergence}

\begin{figure}[!t]
    \centering
    \begin{subfigure}[b]{0.48\linewidth}
        \centering
        \includegraphics[width=\linewidth]{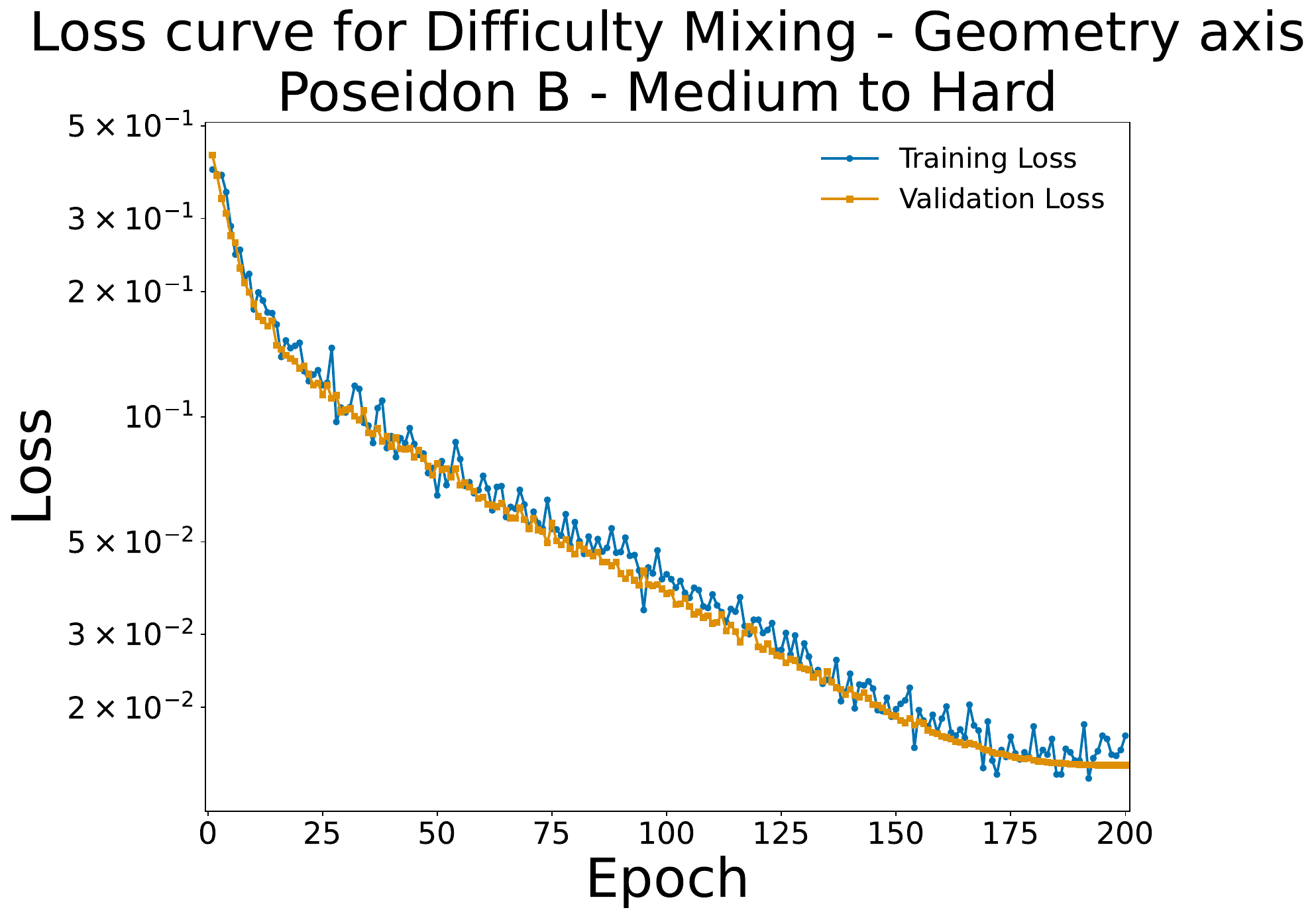}
        \caption{\footnotesize Medium-to-hard geometry.}
        \label{fig:convergence-poseidon-medium-hard}
    \end{subfigure}\hfill
    \begin{subfigure}[b]{0.48\linewidth}
        \centering
        \includegraphics[width=\linewidth]{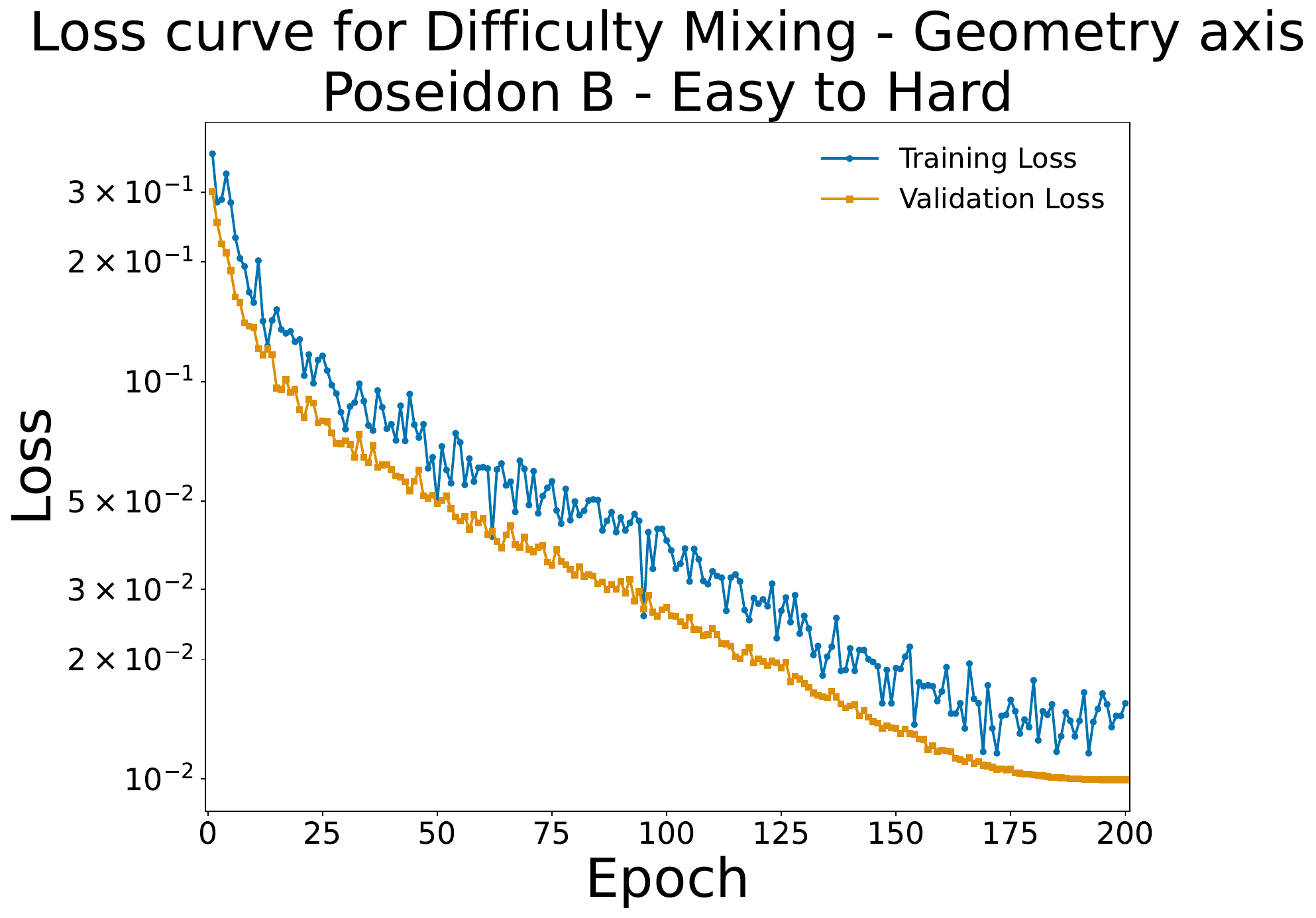}
        \caption{\footnotesize Easy-to-hard geometry.}
        \label{fig:convergence-poseidon-easy-hard}
    \end{subfigure}
    \caption{\textbf{Training convergence for Poseidon-B across difficulty mixing scenarios.} Both subplots show training and validation loss (L1 loss) over epochs for models trained with varying fractions of easy/medium and hard examples. Training loss is computed over the full training set (with mixed difficulty ratio), while validation loss is computed over a fixed set of 100 samples (50 from easy/medium, 50 from hard). Convergence is smooth and stable, demonstrating the robustness of the difficulty mixing strategy.}
    \label{fig:convergence-poseidon}
\end{figure}

To provide additional insight into the training dynamics of our models across different difficulty mixing scenarios, we present the convergence behavior during training and validation in Figure~\ref{fig:convergence-poseidon}. The training loss is computed as the mean L1 loss over the training set, which consists of samples from both easy/medium and hard data according to the difficulty ratio of each experiment. The validation loss is computed as the mean L1 loss over a fixed validation set of 100 samples per experiment: 50 samples drawn from the easy/medium distribution and 50 samples from the hard distribution. Figure~\ref{fig:convergence-poseidon} shows the training and validation loss curves for Poseidon-B across two difficulty mixing scenarios: medium-to-hard and easy-to-hard geometry composition. Both subplots demonstrate smooth convergence and stable validation loss across different mixing ratios, reflecting the robustness of our training procedure and validating the generalization properties of models trained on lower-difficulty data augmented with target-difficulty examples. It is important to note that the validation loss shown here uses a balanced split (50 easy/medium, 50 hard samples) to monitor training stability across difficulty compositions, whereas the test performance numbers reported in the main paper evaluate on the complete hard dataset to assess generalization to the target distribution.

\subsection{Additional Results: FlowBench Harmonics Analysis}\label{app:harmonics}

\begin{figure}[!t]
    \centering
    \includegraphics[width=0.8\linewidth]{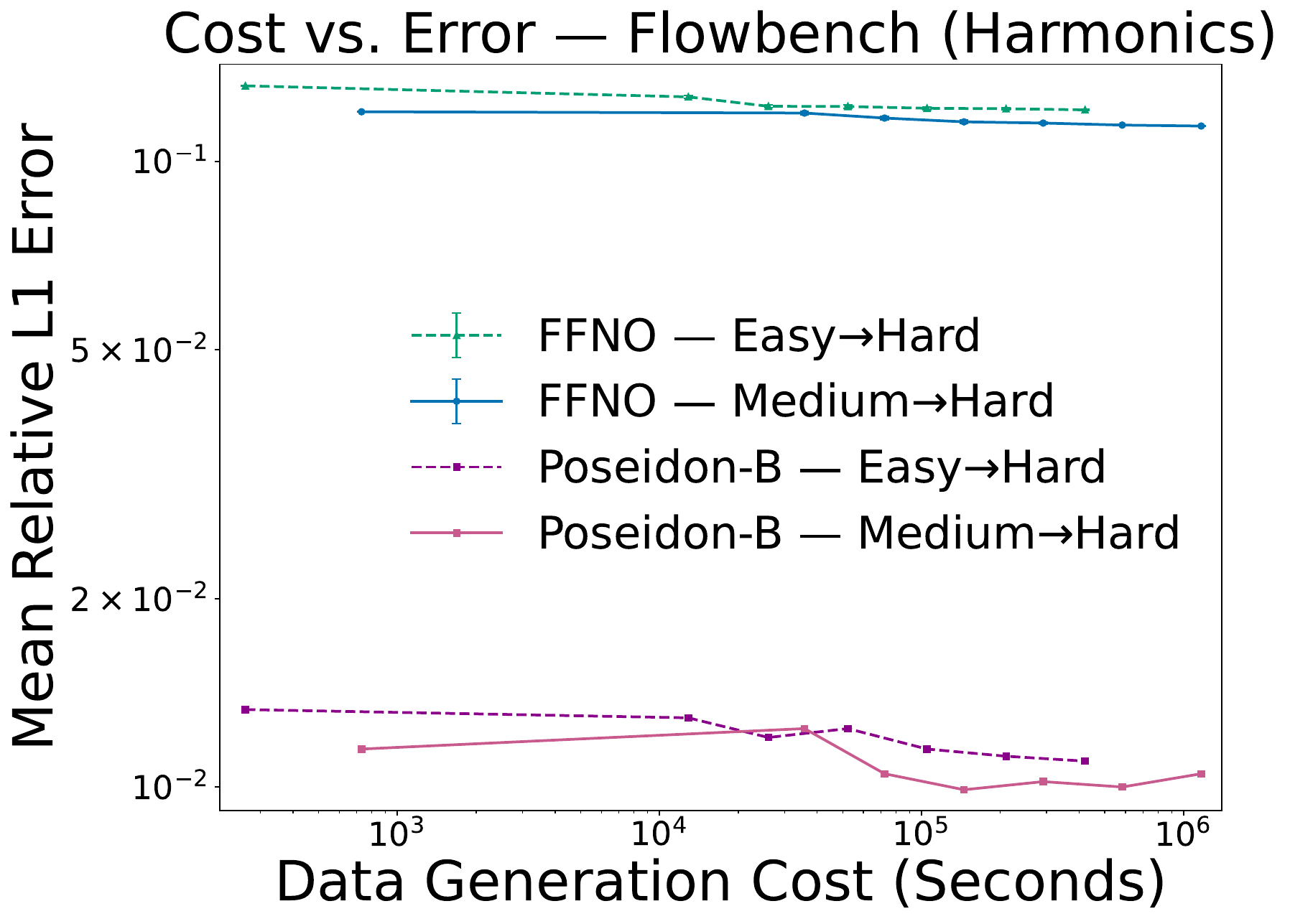}
    \caption{\textbf{Performance on FlowBench's Harmonics FPO simulations} when 100 target examples are augmented with zero-obstacle FPO (easy) or single-obstacle FPO (medium) simulations. Similar to NURBS results, data augmentation with medium-difficulty examples substantially improves performance for most models on the target Harmonics distribution.}
    \label{fig:harmonics-flowbench}
\end{figure}

To further validate the applicability of our findings beyond the primary FPO and LDC domains, we examine the performance on the FlowBench dataset using Harmonics geometries. Similar to the NURBS results presented in the main paper, we augment the target Harmonics examples with zero-obstacle FPO (easy) and single-obstacle FPO (medium) data. Figure~\ref{fig:harmonics-flowbench} presents the cost versus error scaling behavior for models trained on 100 target Harmonics examples augmented with varying data generation costs from lower-difficulty FPO simulations. Consistent with our observations on NURBS geometries and our primary difficulty-mixing experiments, adding lower-difficulty examples substantially reduces error across multiple model architectures. This result demonstrates that difficulty mixing is an effective and generalizable strategy for improving performance on challenging datasets, extending the applicability of our approach across diverse geometric families.

% \section{Technical Appendices and Supplementary Material}
% Technical appendices with additional results, figures, graphs and proofs may be submitted with the paper submission before the full submission deadline (see above), or as a separate PDF in the ZIP file below before the supplementary material deadline. There is no page limit for the technical appendices.

%%%%%%%%%%%%%%%%%%%%%%%%%%%%%%%%%%%%%%%%%%%%%%%%%%%%%%%%%%%%

\end{document}